\documentclass[conference]{IEEEtran}
\IEEEoverridecommandlockouts
\usepackage{cite}
\usepackage{amsmath,amssymb,amsfonts}
\usepackage{algorithmic}
\usepackage{graphicx}
\usepackage{textcomp}
\usepackage{multirow}
\usepackage{amsmath}
\usepackage{mathtools}
\usepackage{csquotes}
\usepackage{enumitem}
\usepackage[table]{xcolor}
\usepackage{booktabs}
\usepackage{authblk}
\usepackage[absolute,overlay]{textpos}
\definecolor{Gray}{gray}{0.85}
\def\BibTeX{{\rm B\kern-.05em{\sc i\kern-.025em b}\kern-.08em
    T\kern-.1667em\lower.7ex\hbox{E}\kern-.125emX}}

\begin{document}
\begin{textblock*}{10cm}(1cm,1cm)
  \footnotesize This paper has been accepted to ICME 2025
\end{textblock*}
\title{Foreground Focus: Enhancing Coherence and Fidelity in Camouflaged Image Generation}

\author[$\,$]{\Large \textit{Pei-Chi Chen$^{1}$, Yi Yao$^{2}$, Chan-Feng Hsu$^{2}$, HongXia Xie$^{3}$, Hung-Jen Chen$^{4}$,}\\ \textit{Hong-Han Shuai$^{2}$, and Wen-Huang Cheng$^{5}$}}

\affil[$\,$]{\normalsize $^{1}$Institute of Information Science, Academia Sinica}
\affil[$2$]{National Yang Ming Chiao Tung University}
\affil[$\,$]{$^{3}$Jilin University, $^{4}$MediaTek, $^{5}$National Taiwan University}
\affil[$\,$]{\texttt{\normalsize pcchen0413@iis.sinica.edu.tw, wenhuang@csie.ntu.edu.tw}}
\maketitle

\begin{abstract}
Camouflaged image generation is emerging as a solution to data scarcity in camouflaged vision perception, offering a cost-effective alternative to data collection and labeling. 
% Previous methods typically blend foreground objects into manually specified backgrounds, increasing labor costs and lacking contextual rationale. 
Recently, the state-of-the-art approach successfully generates camouflaged images using only foreground objects. However, it faces two critical weaknesses: 1) the background knowledge does not integrate effectively with foreground features, resulting in a lack of foreground-background coherence (e.g., color discrepancy); 2) the generation process does not prioritize the fidelity of foreground objects, which leads to distortion, particularly for small objects. To address these issues, we propose a Foreground-Aware Camouflaged Image Generation (FACIG) model. Specifically, we introduce a Foreground-Aware Feature Integration Module (FAFIM) to strengthen the integration between foreground features and background knowledge. In addition, a Foreground-Aware Denoising Loss is designed to enhance foreground reconstruction supervision. Experiments on various datasets show our method outperforms previous methods in overall camouflaged image quality and foreground fidelity. 
\end{abstract}

\begin{IEEEkeywords}
Camouflage image generation, Latent diffusion model
\end{IEEEkeywords}

\section{Introduction}
\label{sec:intro}
Camouflage is a natural adaptation that allows organisms to blend seamlessly into their surroundings. To explore this phenomenon, camouflaged vision perception (e.g., camouflaged object detection \cite{HAITNet}) has gained increasing attention, aiming to detect objects hidden in the environment. This task has broad applications in various fields, including wildlife conservation \cite{wild} and medical \cite{medical}. Despite its significance, the labor-intensive process of collecting and annotating camouflaged images hinders accurate detection and segmentation.

Recent advancements in generative models (e.g., GANs~\cite{gan} and Diffusion models~\cite{ddpm,ldm}) have introduced promising avenues for generating synthetic data to alleviate the scarcity of annotated datasets. 
% For instance, DatasetDM~\cite{datasetdm} employs a text-to-image diffusion model and constructs a unified perception decoder to generate data for various downstream tasks. 
However, most of these methods \cite{bigdatasetgan,datasetdm}  are tailored to generic images, struggling to generalize well to specific domains, e.g. camouflaged images, where maintaining foreground-background texture consistency is crucial.
\begin{figure}[t]
    \centering
    \includegraphics[width=0.42\textwidth]{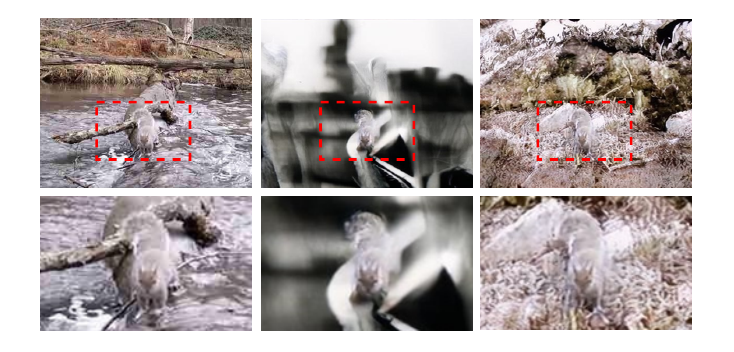}
    \\
    \makebox[0.138\textwidth]{Original image}
    \makebox[0.138\textwidth]{LAKE-RED}
    \makebox[0.138\textwidth]{$\textbf{FACIG (ours)}$}
    \caption{Comparison between the generated images by our method and LAKE-RED~\cite{lake}. The top row shows the complete images, while the bottom row presents zoomed-in views of the highlighted areas. There exists a lack of coherence between foreground and background in LAKE-RED, while our method successfully enhances \textit{foreground-background coherence }and \textit{foreground fidelity}.}
    \label{fig:fig1}

\end{figure}

To better synthesize images of camouflaged objects, several generative models have been proposed, which can be categorized into two classes: \textit{background-guided} and \textit{foreground-guided}. \textit{Background-guided} methods \cite{ci,dci,lcgnet} manually select a background image and then alter the texture details of the given foreground object to match the background. The need for artificially specified backgrounds increases labor costs, and without deliberate selection, the result may lack contextual rationale (e.g., an octopus in the sky). To address this issue, LAKE-RED~\cite{lake} first proposes a \textit{foreground-guided} method, which is based on the Latent Diffusion Models (LDMs)~\cite{ldm} and VQ-VAE~\cite{vqvae}, to generate camouflaged images using only foreground objects. This method is inspired by the high correlation between the foreground features and the background features in camouflaged images. By leveraging foreground features to retrieve background information through a pre-trained codebook, it successfully produces camouflaged images. 

However, Fig.~\ref{fig:fig1} illustrates two principal challenges in the quality of generated camouflaged images: 1) \textit{visual inconsistency} (e.g., color discrepancy) between the generated background scenes and the foreground objects, and 2) \textit{distortion} in the reconstructed foreground objects, notably in smaller objects. These issues stem from two main limitations in the current models. 
% Firstly, the background generation process is solely dependent on retrieved background knowledge, which fails to incorporate critical foreground details. This oversight results in backgrounds that may not align visually with the foreground objects, undermining the camouflage effect. 
Firstly, the background retrieval process only ensures a basic level of foreground-background compatibility, focusing on feature alignment rather than deep, interactive fusion. This oversight results in texture and visual inconsistency between the foreground and background, weakening the camouflage effect. 
Secondly, the denoising process within LDMs focuses on reconstructing the entire latent feature, neglecting the priority of foreground fidelity. This approach tends to blur or distort foreground objects, especially those with finer details, thereby diminishing the image quality.
% effectiveness of the camouflage. 

% Motivated by these reasons, we propose a foreground-aware camouflaged image generation (FACIG) method to mitigate these issues. 
To mitigate the challenges mentioned above, we propose a foreground-aware camouflaged image generation (FACIG) model. We introduce a foreground-aware feature integration module (FAFIM) to effectively strengthen the integration between the foreground features and the retrieved background knowledge, bridging the texture gap between the generated background scenes and foreground objects. Besides, we design a foreground-aware denoising loss to enhance foreground reconstruction supervision, resulting in clearer foreground.

Our contributions are summarized as follows:
\begin{itemize}
    \item To address the inconsistency between background and foreground, we introduce a foreground-aware feature integration module (FAFIM) that effectively integrates the foreground features into the background knowledge.
    \item To maintain foreground fidelity, we design a foreground-aware denoising loss that emphasizes accurate foreground reconstruction during the denoising process.
    \item Extensive experimental results demonstrate the effectiveness of FACIG in generating high-quality camouflaged images, outperforming state-of-the-art (SOTA) methods.
\end{itemize}
\section{Related Work}
\subsection{Synthetic Dataset Generation} 
Synthetic data has been a cost-effective solution to data scarcity in deep learning methods. Recently, the evolution of generative models (e.g., GANs and Diffusion Models) has further expanded the potential of synthetic data. BigDatasetGAN~\cite{bigdatasetgan} exploits the feature space of a trained GAN to produce data with pixel-level annotations. 
% Based on DDPM~\cite{ddpm}, DatasetDDPM~\cite{datasetddpm} uses feature maps of U-Net to generate semantic segmentation data.
% DiffuMask~\cite{diffumask} exploits the potential of the cross-attention map between text and image from the LDM to generate images with semantic mask annotation. 
DatasetDM~\cite{datasetdm} employs a text-to-image diffusion model and constructs a unified perception decoder to generate data for various downstream tasks like segmentation, depth estimation, and pose estimation. However, these methods concentrate on generic image generation, failing to generalize to camouflaged images.

\subsection{Camouflaged Image Generation}
Camouflage image generation methods can be broadly categorized into two main classes: background-guided and foreground-guided approaches. The background-guided methods blend the foreground objects into the background images by altering their texture and style. 
% The first approach of this kind was introduced by~\cite{ci}, which uses hand-crafted features to replace texture details from the foreground with those from the background, effectively concealing the objects within the background. 
% A handcrafted feature replaces texture details from the foreground with those from the background, effectively concealing the objects in the background~\cite{ci}.
% With the advancements of deep learning,~\cite{dci} proposes a deep camouflage generation approach and designs an attention-aware camouflage loss to preserve only salient features of concealed objects. 
LCG-Net~\cite{lcgnet} adaptively combines the features of the foreground and background images by point-to-point structure similarity to hide the objects in regions with multiple appearances. While these methods successfully conceal objects within background images, they require manual specification of backgrounds, increasing labor costs. Motivated by this, LAKE-RED~\cite{lake} first proposes a foreground-guided method and utilizes foreground features to retrieve background information through a pre-trained codebook. 
% However, texture inconsistency between the foreground and background, combined with foreground distortion, degrades the quality of the generated images.
% Despite this, texture inconsistency between foreground and background, coupled with foreground distortion, leads to poor image quality.
However, foreground distortion and texture inconsistency lead to poor image quality.
\section{Methodology}
\subsection{Preliminary: Latent Diffusion Models}
Our method leverages Latent Diffusion Models (LDMs) \cite{ldm}, which perform the denoising process in latent space to decrease the computational costs of high-resolution image generation.  Specifically, given an image $\mathbf{x_0}$, a pre-trained autoencoder $\mathcal{E}$ is first utilized to encode $\mathbf{x_0}$ into a low-dimensional latent representation $\mathbf{z_0}$. Then the forward process gradually adds Gaussian noise to $\mathbf{z_0}$. The latent $\mathbf{z}$ at time step $t$ can be expressed as:
\begin{equation}
\mathbf{z_t}=\sqrt{\bar{\alpha}_t} \mathbf{z_0} +\sqrt{1-\bar{\alpha}_t}\epsilon, \quad\epsilon\sim\mathcal{N}(0,\mathbf{I}),
\end{equation}
where $t\in \{1,...,T\}$, $T$ represents the number of time step in forward process, and $\bar{\alpha}_t$ is a variance schedule. The reverse process employs a conditional UNet $\epsilon_\theta$ to predict the noise added on $\mathbf{z_0}$. The training objectiveness is defined as:
\begin{equation}
\mathcal{L}=\mathbb{E}_{\mathbf{z},t,\epsilon}\Vert\epsilon-\epsilon_\theta(\mathbf{z_t},\mathbf{c},t)\Vert^2_2,
\label{eq:ldm}
\end{equation}
where $\mathbf{c}$ is the condition guiding the generation process. 

After $T$ steps of the reverse process, the noisy latent is recovered to $\mathbf{z^\prime_0}$ with the noise removed. Finally, in this work, we use a VQ-VAE-based decoder $\mathcal{D}$ to reconstruct the image from this latent representation. The visual information from a codebook $\mathbf{e}$ is integrated into the latent representation through a quantization layer $\mathbf{\nu}$ within the decoder. The decoding process is expressed as:
\begin{equation}
    \mathbf{x^\prime_0}=\mathcal{D}(\nu(\mathbf{e},\mathbf{z^\prime_0})),
\end{equation}
where $\mathbf{e} \in \mathbb{R}^{K\times D}$, with K and D denoting the number and dimension of latent embedding vectors, respectively.

\subsection{Model Overview}
\begin{figure*}
\centering
\includegraphics[height=!,width=0.88\linewidth,keepaspectratio=true]{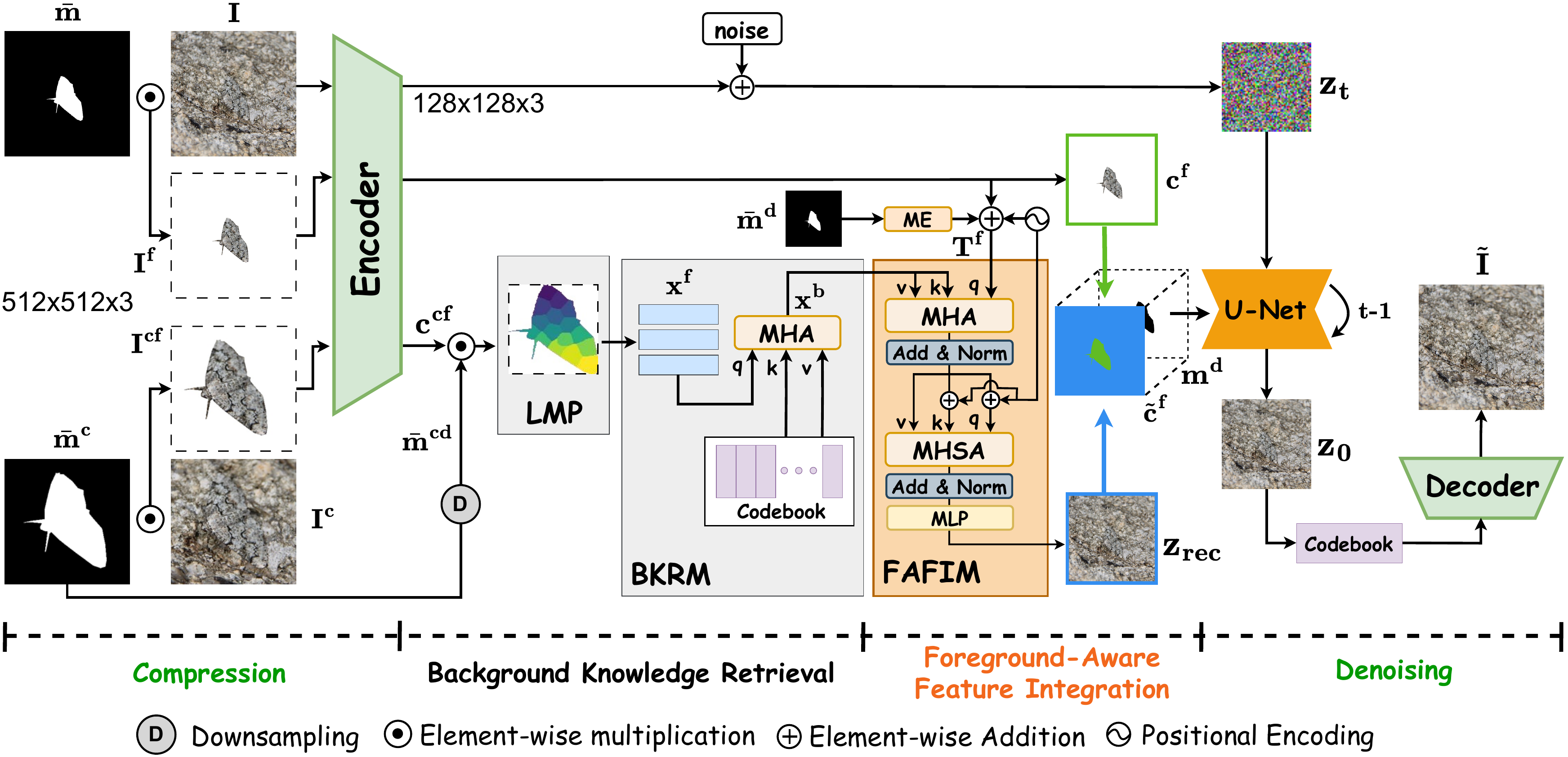}
\caption{Overall framework of FACIG, which comprises four stages: (1) compression of the input image into a latent space. (2) background knowledge retrieval from the codebook. (3) foreground-aware feature integration, where foreground features are integrated into the background knowledge, and (4) denoising process guided by the integrated features.}
\label{fig:facig}
\end{figure*}

The overall framework of our model is shown in Fig. \ref{fig:facig}. Our model builds upon LAKE-RED~\cite{lake}, which is inspired by image inpainting techniques and leverages LDM to generate a background scene, conditioned by the information derived from the foreground. Formally, given a source image $\textbf{I}^{s}$ and an object mask $\textbf{m}^{s}$, we first resize both to match the model's input size. This results in an input image $\textbf{I} \in \mathbb{R}^{H\times W\times 3}$, and a binary mask \textbf{m} $\in \{0,1\}^{H\times W}$ which indicates the object's location. Specifically, $m_{i,j}=0$, with $i \in \{1,2,\ldots,H\}$ and $j \in \{1,2,\ldots,W\}$, represents the object region to be maintained and $m_{i,j}=1$ represents the background region to be edited. The model takes $\{\textbf{I},\textbf{m}\}$ as input and outputs a camouflaged image $\tilde{\textbf{I}}$. The condition \textbf{c} consists of a mask $\textbf{m}^d$ indicating the editable background regions, and a feature $\tilde{\textbf{c}}^f$ guiding the generation. The condition can be defined as:
\begin{equation}
\begin{gathered}
    \textbf{c} = Concat\left( \tilde{\textbf{c}}^f, \textbf{m}^d \right),\\
    \tilde{\textbf{c}}^f=\textbf{c}^f\cdot\left(1-\textbf{m}^d\right)+\textbf{z}_{rec}\cdot\textbf{m}^d, \quad \textbf{c}^f=\mathcal{E}\left(\textbf{I}^{f}\right),
\end{gathered}
\end{equation}
where $\textbf{I}^{f}=\textbf{I}\odot \bar{\textbf{m}}$ with $\bar{\textbf{m}}=1-\textbf{m}$, $\textbf{c}^f \in \mathbb{R}^{h\times w\times 3}$, and $\textbf{m}^d$ is obtained by downsampling $\textbf{m}$ with a factor $f=2^n$ to fit the required shape. $\textbf{Z}_{rec}$ denotes the background features we aim to construct. Thus, $\tilde{\textbf{c}}^f$ can be interpreted as $\textbf{c}^f$ with the background regions filled by $\textbf{Z}_{rec}$ and the foreground regions preserved.

The overall process of constructing $\textbf{Z}_{rec}$ is as follows: to avoid losing foreground information due to resizing, we first crop the foreground region from the source image $\textbf{I}^s$. Since the cropped object is typically smaller than the model's input size, we pad it to match the required dimensions, resulting in $\textbf{I}^c$ and the corresponding mask $\bar{\textbf{m}}^c$. We then derive the cropped foreground features $\textbf{c}^{cf}$ by encoding the cropped foreground image $\textbf{I}^{cf}$, where $\textbf{I}^{cf}=\textbf{I}^c\odot\bar{\textbf{m}}^c$. Following LAKE-RED, we use Localized Masked Pooling (LMP) and the Background Knowledge Retrieval Module (BKRM) to retrieve rich background knowledge. Next, we introduce the Foreground-Aware Feature Integration Module (FAFIM) to effectively integrate the foreground features and background knowledge to construct $\textbf{Z}_{rec}$. Further details of the model design are provided in the following sections.

\subsection{Background Knowledge Retrieval}
Since using only foreground features to generate diverse background scenes is insufficient, we leverage the pre-trained codebook $\textbf{e}$ from VQ-VAE to provide rich background knowledge. Specifically, foreground features $\textbf{x}^f$ are used to retrieve background features $\textbf{x}^b$ from $\textbf{e}$. Note that the expressiveness of the feature representation $\textbf{x}^f$ extracted from $\textbf{c}^{cf}$ significantly influences the richness of the background knowledge that can be retrieved from the codebook. 
To extract representative foreground features, we employ SLIC algorithm~\cite{slic} to cluster the foreground regions into $S$ superpixels. The representative foreground feature vectors are then obtained by an averaging function $\Phi$ that averages the features within each superpixel. This process can be expressed as:
\begin{equation}
\begin{gathered}    \mathbf{p}_{i,1},\mathbf{p}_{i,2},\cdots,\mathbf{p}_{i,S}=SLIC\left(\mathbf{c}_i^f,\mathbf{m}^{cd}\right),\\
x_{i,j}^{f}=\Phi\left(\mathbf{p}_{i,j},\mathbf{c}_i^f\right)=\frac{\Sigma^{w}_{x=1}\Sigma^{h}_{y=1}\,c^f_{i,(x,y)}\cdot p_{i,j,(x,y)}}{\Sigma^{w}_{x=1}\Sigma^{h}_{y=1}\,p_{i,j,(x,y)}},
\end{gathered}
\end{equation}
where $\bar{\textbf{m}}^{cd}$ is the downsampled $\bar{\textbf{m}}^c$. $i$ is the channel index and $j$ is the superpixel index. This yields $\textbf{x}^f \in \mathbb{R}^{S\times 3}$. 

The retrieval process is operated by a multi-head attention ($MHA$) layer with $H$ heads, which can be formulated as: 
\begin{equation}
    \textbf{x}^b = MHA\left(\textbf{x}^f,\textbf{e}^T,\textbf{e}^T\right), 
\end{equation}
where $\textbf{e}^T$ is the transpose of $\textbf{e}$, and
\begin{equation}
\begin{split}
MHA\left(\textbf{Q},\textbf{K},\textbf{V}\right)=Concat\left(\textbf{h}_1,\textbf{h}_2,\ldots,\textbf{h}_H\right)\textbf{W}^O,\hspace{0.5cm} \\ 
\textbf{h}_i\left(\textbf{Q},\textbf{K},\textbf{V}\right)= Softmax\left(\frac{\left(\textbf{Q}\textbf{W}_i^Q\right)\left(\textbf{K}\textbf{W}_i^K\right)^T}{\sqrt{d_k}}\right)\textbf{V}\textbf{W}_i^V,
\end{split}
\end{equation}
with $\textbf{W}^O, \textbf{W}^Q,\textbf{W}^K,\textbf{W}^V$ the learnable weights, and $d_k$ the channel dimension.
\subsection{Foreground-Aware Feature Integration Module}
To generate a background scene that better harmonizes with the foreground object, we introduce a Foreground-Aware Feature Integration Module (FAFIM) for integrating the retrieved background information $\textbf{x}^b$ with the foreground features $\textbf{c}^f$. Specifically, we first add mask embeddings to $\textbf{c}^f$ to provide region information and reshape it into a sequence of patches $\textbf{x}^p \in \mathbb{R}^{N\times P^2\times3}$, where $N=hw/P^2$, $\left(P,P\right)$ is the resolution of each patch. These patches are then mapped to foreground tokens $\textbf{T}^f \in \mathbb{R}^{N\times C}$ using a convolution projector, with positional encoding ($PE$) added to further enhance the awareness of the locality relationship between the foreground and background. The process can be summarized as:
\begin{equation}
    \textbf{T}^f = Conv\left(ME\left(\bar{\textbf{m}}^d\right)+\textbf{c}^f\right)+PE,
\end{equation}
where $ME\left(\cdot\right)$ is a mask embedding look-up table.
% mapping $\bar{\textbf{m}}^d$ to the mask embeddings. 

The integration process begins by embedding the background knowledge $\textbf{x}^b$ into each token of $\textbf{T}^f$ using an $MHA$ layer, along with a residual connection and layer normalization, which can be formulated as:
\begin{equation}
\textbf{T}^{fb}=LN\left(MHA\left(\textbf{T}^f, \textbf{x}^b,\textbf{x}^b\right)+\textbf{T}^f\right).
\end{equation}
The resulting features $\textbf{T}^{fb}$ contain both foreground features and background knowledge. However, since $\textbf{c}^f$ lacks information in the background regions, the foreground-background integration only operates in the foreground regions of $\textbf{T}^f$, which will not be used to construct the condition $\tilde{\textbf{c}}^f$. The background regions of $\textbf{T}^f$ only contain the background knowledge from $\textbf{x}^b$. To address this issue, we use a multi-head self-attention (MHSA) layer to embed the foreground features into the background regions. The formula can be written as:
\begin{equation}
\begin{gathered}
\textbf{T}^{integ}=LN\left(MHSA\left(\textbf{T}^{fb}\right)+\textbf{T}^{fb}\right), \\
MHSA\left(\textbf{T}^{fb}\right)=MHA\left(\textbf{T}^{fbP},\textbf{T}^{fbP},\textbf{T}^{fb}\right), 
\end{gathered}
\end{equation}
where $\textbf{T}^{fbP}=\textbf{T}^{fb}+PE$. By the $MHSA$ layer, the foreground features and the background knowledge interact with each other, letting the background regions be aware of the foreground features and encouraging consistency between the generated background scene and the foreground. Finally, we upsample the integrated tokens $\textbf{T}^{integ}$ and use an MLP layer to reconstruct the GT image features $\textbf{z}_0=\mathcal{E}\left(\textbf{I}\right)$, $\textbf{z}_0 \in \mathbb{R}^{h\times w\times 3}$. The reconstruction features can be computed as:
\begin{equation}
\textbf{z}_{rec}=MLP\left(upsample\left(\textbf{T}^{integ}\right)\right).
\end{equation}
\subsection{Loss Function}
$\bullet$ \textbf{Foreground-Aware Denoising Loss.} The standard loss function (Eq. \ref{eq:ldm}) in LDMs minimizes the difference between the predicted noise and the added noise across the whole latent $\mathbf{z_0}$. However, in cases where the foreground object occupies a relatively small portion of the image, the loss function may result in the model achieving a low overall loss by primarily reconstructing the background accurately, thereby neglecting the detailed reconstruction of the foreground object. Motivated by this, we reformulate the loss function as:
\begin{equation} 
\begin{split}
\mathcal{L}_{FADL} & = \mathbb{E}_{\mathbf{z},t,\epsilon}\Vert w\cdot \mathbf{m}^d\cdot(\epsilon-\epsilon_\theta(\mathbf{z_t},\mathbf{c},t))\Vert^2_2 \\
& + \mathbb{E}_{\mathbf{z},t,\epsilon}\Vert (1-\mathbf{m}^d)\cdot(\epsilon-\epsilon_\theta(\mathbf{z_t},\mathbf{c},t))\Vert^2_2,
\end{split}
\label{eq:FADL}
\end{equation}
which comprises two components: one for the foreground and one for the background. To emphasize the foreground region, we introduce a weighting factor $w$ to upweight the foreground component. $w$ is designed to be negatively correlated to the portion of the foreground object in the image. Thus, the smaller the foreground object is, the larger $w$ becomes, and vice versa. Specifically, we define $w$ as:
\begin{equation}
    w=\frac{1}{\alpha + r},\,\,\,\,r=\frac{ForegroundArea}{H*W},
\end{equation}
where $r$ represents the ratio of the foreground area to the image. To prevent huge values of 
$w$ caused by tiny objects, which can lead to harmful gradients and destabilize training, we introduce the regularization term $\alpha$ to limit the weight's upper bound.  In our experiments, we set $\alpha$ to 0.125, resulting in an upper bound of 8 for weight.
\\
$\bullet$ \textbf{Overall Loss.}
The overall loss can be formulated as:
\begin{equation}
    \mathcal{L} = \lambda\mathcal{L}_{FADL} + \mathcal{L}_{bgrec},
\end{equation}
where $\lambda$ is the hyperparameter to balance the contributions of the two losses, and $\mathcal{L}_{bgrec}$ is the loss of background latent reconstruction, which can be written as:
\begin{equation}
\mathcal{L}_{bgrec}=\mathbb{E}_{\mathbf{z}}\Vert (1-\mathbf{m}^d)\cdot(\mathbf{z_{rec}}-\mathbf{z_0})\Vert^2.
\end{equation}

\begin{table*}[t]
\centering
\tabcolsep=7.5pt
\renewcommand\arraystretch{0.85}
\caption{Quantitative overall camouflaged image quality results of our method and state-of-the-art (SOTA) image generation methods across three subsets. $\dagger$ indicates camouflaged image generation methods.}
\begin{tabular}{l|cccccccc}
\toprule[1.5pt]
\multicolumn{1}{c|}{\multirow{2.5}{*}{Methods}} &\multicolumn{2}{c}{Camouflaged Objects} &\multicolumn{2}{c}{Salient Objects} &\multicolumn{2}{c}{General Objects} &\multicolumn{2}{c}{Overall}\\ \cmidrule(rl){2-3} \cmidrule(rl){4-5} \cmidrule(rl){6-7} \cmidrule(rl){8-9} 
&  FID $\downarrow$ & KID $\downarrow$ &  FID $\downarrow$ & KID $\downarrow$ &  FID $\downarrow$ & KID $\downarrow$ &  FID $\downarrow$ & KID $\downarrow$\\
\midrule[1pt] 
\multicolumn{9}{c}{\textbf{Background-Guided / Image Blending}} \\
\midrule[1pt]
% $\text{AB}_{\,IPOL'16}$  &117.11 &0.0645& 126.78 &0.0614 &133.89 &0.0645 &120.21 &0.0623 \\
% $\text{AdaIN}_{\,ICCV'17}$ &125.16 &0.0721 &133.20 &0.0702 &136.93 &0.0714 &126.94 &0.0703 \\
$\text{CI}_{\,TOG'10}$ $\dagger$ &124.49 &0.0662 &136.30 &0.0738 &137.19 &0.0713 &128.51 &0.0693 \\
$\text{DCI}_{\,AAAI'20}$ $\dagger$ &130.21 &0.0689 &134.92 &0.0665 &137.99 &0.0690 &130.52 &0.0673 \\
$\text{LCGNet}_{\,TMM'22}$ $\dagger$ &129.80 &0.0504 &136.24 &0.0597 &132.64 &0.0548 &129.88 &0.0550\\
\midrule[1pt] 
\multicolumn{9}{c}{\textbf{Foreground-Guided / Image Inpainting}} \\
\midrule[1pt]
$\text{TFill}_{\,CVPR'22}$ &63.74 &0.0336 &96.91 &0.0453 &122.44 &0.0747 &80.39 &0.0438 \\
$\text{LDM}_{\,CVPR'22}$  &58.65 &0.0380 &107.38 &0.0524 &129.04 &0.0748 &84.48 &0.0488 \\
$\text{RePaint-L}_{\,CVPR'22}$ &76.80 &0.0459 &114.96 &0.0497 &136.18 &0.0686 &96.14 &0.0498 \\
$\text{LAKE-RED}_{\,CVPR'24}$ $\dagger$  & 39.55 &0.0212 &88.70 &0.0428 &102.67 &0.0625 &64.27 &0.0355\\
\rowcolor{Gray}
\textbf{FACIG (Ours)}  &\textbf{27.61} &\textbf{0.0099} &\textbf{82.23}	&\textbf{0.0326} &\textbf{96.94} &\textbf{0.0503} &\textbf{52.87} &\textbf{0.0229}\\						
\midrule[1.5pt] 
\end{tabular}
\label{table:quality}
\end{table*}
\section{Experiments}
\subsection{Experimental settings}
\noindent $\bullet$ \textbf{Datasets.}
% Following LAKE-RED~\cite{lake}, we utilize 4040 real images (3040 from COD10K~\cite{COD10K} and 1000 from CAMO~\cite{CAMO}) for training our model. In evaluation, we collect three subsets of data as testing data: Camouflaged Objects (CO), Salient Objects (SO), and General Objects (GO). The CO subset comprises 6473 images from the COD10K, CAMO, and NC4K datasets~\cite{NC4K}. For the SO and GO subsets, we employ the datasets collected by LAKE-RED. SO and GO subsets are used to evaluate the model's capability for transferring images from non-camouflaged to camouflaged.
Following LAKE-RED~\cite{lake}, we train our model on 4040 real images (3040 from COD10K~\cite{COD10K} and 1000 from CAMO~\cite{CAMO}). For evaluation, we use three subsets: Camouflaged Objects (CO), Salient Objects (SO), and General Objects (GO)~\footnote{Due to space constraints, more details can be found in the appendix.}. 
% The CO subset includes 6473 images from COD10K, CAMO, and NC4K~\cite{NC4K}, while SO and GO subsets, collected by LAKE-RED, assess the model’s ability to transfer non-camouflaged to camouflaged images.

\noindent $\bullet$ \textbf{Evaluation metrics.}
% We use Fréchet Inception Distance (FID)~\cite{fid} and Kernel Inception Distance (KID)~\cite{kid} metrics to measure the quality of generated camouflaged images by calculating the distribution discrepancy compared to real camouflaged images. To assess the foreground fidelity, peak signal-to-noise ratio (PSNR) and structural similarity index measure (SSIM) on foreground objects are measured by masking out the background regions using object masks.\\
We evaluate the quality of generated images using Fréchet Inception Distance (FID)\cite{fid} and Kernel Inception Distance (KID)\cite{kid}, which measure distribution discrepancies compared to real camouflaged images. For foreground fidelity, we calculate peak signal-to-noise ratio (PSNR) and structural similarity index measure (SSIM) on foreground objects, masking out background regions with object masks.\\
\noindent $\bullet$ \textbf{Baselines.}
% We compare our method with existing camouflaged image generation methods, such as CI \cite{ci}, DCI \cite{dci}, LCGNet \cite{lcgnet}, and LAKE-RED \cite{lake}. Besides, we compare with state-of-the-art generic image generation methods, such as  TFill \cite{tfill}, LDM \cite{ldm}, and RePaint-L \cite{repaint}.
We compare our method with camouflaged image generation approaches (CI~\cite{ci}, DCI~\cite{dci}, LCGNet~\cite{lcgnet}, LAKE-RED~\cite{lake}) and state-of-the-art generic methods (TFill~\cite{tfill}, LDM~\cite{ldm}, RePaint-L~\cite{repaint}).
% \noindent $\bullet$ \textbf{Implementation Details.}
% We implement our model based on PyTorch and NVIDIA Tesla V100 GPUs. Our method employs LDM~\cite{ldm} pre-trained on inpainting tasks for initialization and a pre-trained VQ-VAE~\cite{vqvae} as the autoencoder. During training, the autoencoder and the codebook are frozen. The model is trained with 50K iterations and a batch size of 8. Other hyperparameters are set similar to LDM.
% We implement our model using PyTorch on NVIDIA Tesla V100 GPUs. The model is initialized with a pre-trained LDM and incorporates a pre-trained VQ-VAE~\cite{vqvae} as the autoencoder. During training, the autoencoder and codebook remain frozen. The model undergoes 50K iterations with a batch size of 8, and other hyperparameters are set similarly to those in LDM.

\subsection{Performance Analysis}
% We provide quantitative and qualitative comparisons of our method with various state-of-the-art techniques, not limited to camouflaged image generation. In addition, recognizing that evaluation metrics may not fully capture the quality of the generated images, we conduct a user study to assess human preferences. Finally, ablation studies are performed to verify the effectiveness of each proposed component.
\begin{table}[t]
    \centering
    \tabcolsep=1.3pt
    \renewcommand\arraystretch{0.9}
    \caption{Quantitative foreground reconstruction results of our method and LAKE-RED. \enquote{(f)} refers to the full dataset, and \enquote{(s)} refers to small objects.}
    \begin{tabular}{@{}>{\centering}p{1.2cm}|l|cccc@{}}
    \toprule[1.5pt]
    Dataset &\multicolumn{1}{c|}{Methods} &PSNR (f)$\uparrow$ &PSNR (s)$\uparrow$ &SSIM (f)$\uparrow$ &SSIM (s)$\uparrow$ \\ 
    \midrule[1pt]
   Camou. &LAKE-RED &18.09 &14.39 &0.705 &0.391\\
   \rowcolor{Gray}
    \cellcolor{white} Objects &\textbf{FACIG (Ours)}  &\textbf{20.80} &\textbf{16.86} &\textbf{0.808} &\textbf{0.572}\\
    \midrule[1pt]
    Salient &LAKE-RED &19.79	&14.57 &0.809 &0.472\\
    \rowcolor{Gray}
   \cellcolor{white} Objects &\textbf{FACIG (Ours)}  &\textbf{26.06} &\textbf{20.49} &\textbf{0.943} &\textbf{0.805}\\
    \midrule[1pt]
    General &LAKE-RED &22.21	&16.88 &0.870 &0.584\\
    \rowcolor{Gray}
   \cellcolor{white} Objects &\textbf{FACIG (Ours)}  &\textbf{23.50} &\textbf{17.85} &\textbf{0.906} &\textbf{0.716}\\
    \midrule[1.5pt] 
    \end{tabular}
\label{table:foreground}
\vspace{-0.5cm}
\end{table}

\begin{figure*}[t]
    \centering
    \raisebox{0.7\height}{\makebox[0.14\textwidth][]{\shortstack[c]{Camouflaged \\Objects}}}
    \includegraphics[width=0.09\textwidth,height=0.09\textwidth]{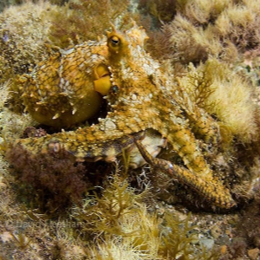}
    \hspace{-3pt} \includegraphics[width=0.09\textwidth,height=0.09\textwidth]{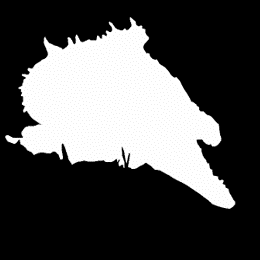}
    \hspace{-3pt} \includegraphics[width=0.09\textwidth,height=0.09\textwidth]{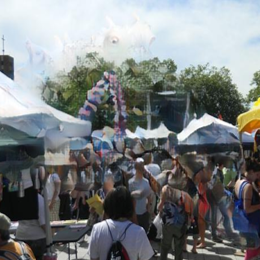}
    \hspace{-3pt} \includegraphics[width=0.09\textwidth,height=0.09\textwidth]{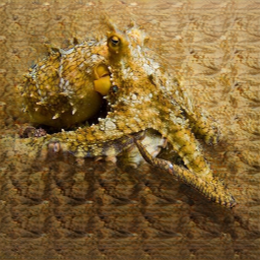}
    \hspace{-3pt} \includegraphics[width=0.09\textwidth,height=0.09\textwidth]{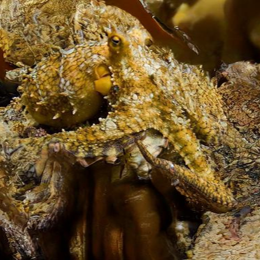}
    \hspace{-3pt} \includegraphics[width=0.09\textwidth,height=0.09\textwidth]{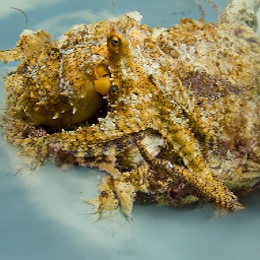}
    \hspace{-3pt} \includegraphics[width=0.09\textwidth,height=0.09\textwidth]{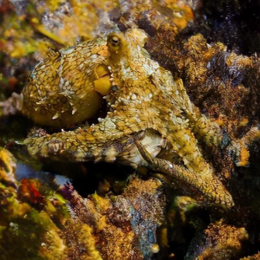}
    \hspace{-3pt} \includegraphics[width=0.09\textwidth,height=0.09\textwidth]{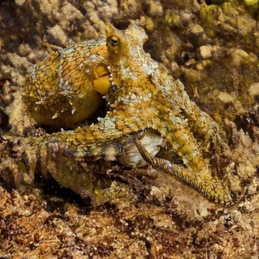}
    \makebox[0.07\textwidth][]{}
    \vspace{1pt}
    \\
    \raisebox{0.8\height}{\makebox[0.14\textwidth]{\shortstack[c]{Salient \\Objects}}}
    \includegraphics[width=0.09\textwidth,height=0.09\textwidth]{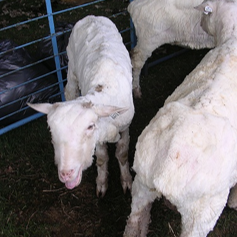}
    \hspace{-3pt} \includegraphics[width=0.09\textwidth,height=0.09\textwidth]{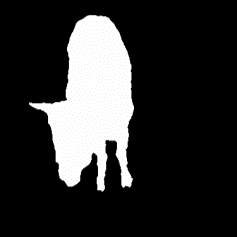}
    \hspace{-3pt} \includegraphics[width=0.09\textwidth,height=0.09\textwidth]{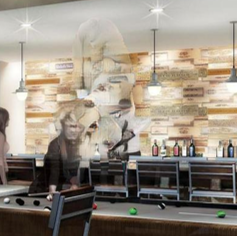}
    \hspace{-3pt} \includegraphics[width=0.09\textwidth,height=0.09\textwidth]{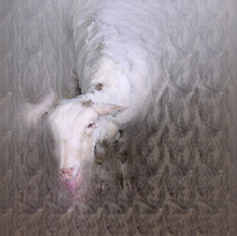}
    \hspace{-3pt} \includegraphics[width=0.09\textwidth,height=0.09\textwidth]{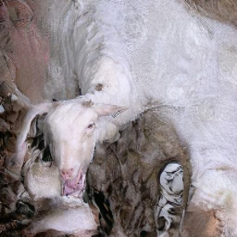}
    \hspace{-3pt} \includegraphics[width=0.09\textwidth,height=0.09\textwidth]{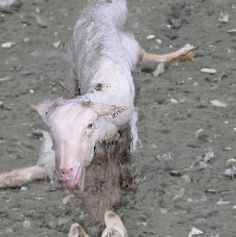}
    \hspace{-3pt} \includegraphics[width=0.09\textwidth,height=0.09\textwidth]{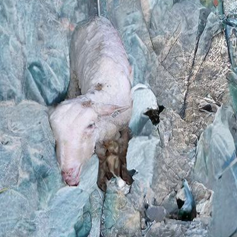}
    \hspace{-3pt} \includegraphics[width=0.09\textwidth,height=0.09\textwidth]{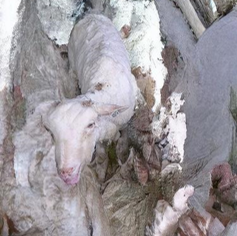}
    \makebox[0.07\textwidth][]{}
    \vspace{1pt}
    \\
    \raisebox{0.8\height}{\makebox[0.14\textwidth]{\shortstack[c]{General \\Objects}}}
    \includegraphics[width=0.09\textwidth,height=0.09\textwidth]{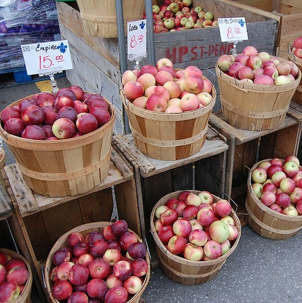}
    \hspace{-3pt} \includegraphics[width=0.09\textwidth,height=0.09\textwidth]{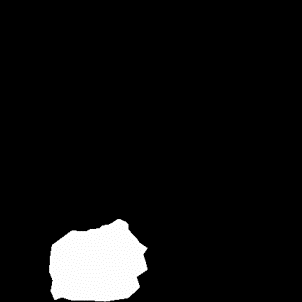}
    \hspace{-3pt} \includegraphics[width=0.09\textwidth,height=0.09\textwidth]{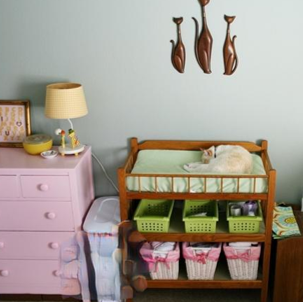}
    \hspace{-3pt} \includegraphics[width=0.09\textwidth,height=0.09\textwidth]{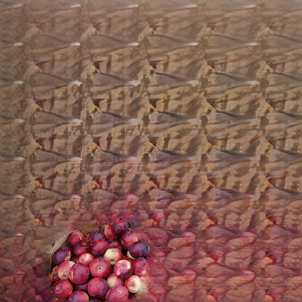}
    \hspace{-3pt} \includegraphics[width=0.09\textwidth,height=0.09\textwidth]{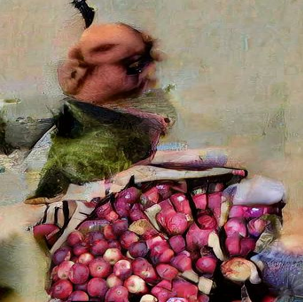}
    \hspace{-3pt} \includegraphics[width=0.09\textwidth,height=0.09\textwidth]{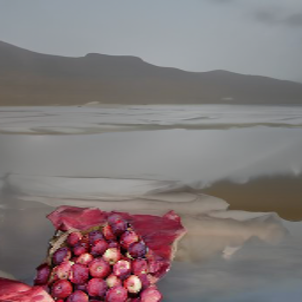}
    \hspace{-3pt} \includegraphics[width=0.09\textwidth,height=0.09\textwidth]{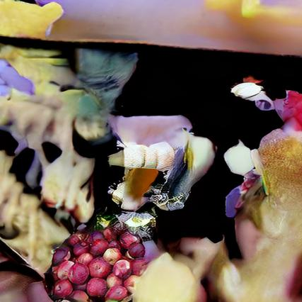}
    \hspace{-3pt} \includegraphics[width=0.09\textwidth,height=0.09\textwidth]{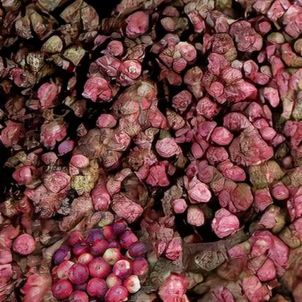}
    \makebox[0.07\textwidth][]{}
    \vspace{2pt}
    \\
    \makebox[0.14\textwidth]{}
    \makebox[0.09\textwidth]{Image}
    \makebox[0.09\textwidth]{Mask}
    \makebox[0.09\textwidth]{LCGNet}
    \makebox[0.09\textwidth]{TFill}
    \makebox[0.09\textwidth]{LDM}
    \hspace{-1pt}\makebox[0.09\textwidth]{RePaint-L}
    \makebox[0.09\textwidth]{LAKE-RED}
    \makebox[0.09\textwidth]{\textbf{Ours}}
    \makebox[0.07\textwidth][]{}
    \caption{Qualitative results of our method and SOTA image generation methods. The first two columns are the original images and the object masks.}
    \label{fig:visualization}
\end{figure*}
\noindent $\bullet$ \textbf{Camouflaged Image Quality.}
Tab. \ref{table:quality} presents the quantitative results on overall camouflaged image quality. Background-guided methods yield higher FID and KID scores due to domain gaps between manually specified backgrounds and real camouflaged images. Foreground-guided methods generate better background scenes from the objects. However, most of these methods are designed for generic image generation, leading to distribution discrepancies. In contrast, our FACIG achieves the best performance, improving FID by 17.7$\%$ and KID by 35.5$\%$ over the previous SOTA. This highlights FACIG's effectiveness in generating high-quality, realistic camouflaged images with improved foreground-background coherence.
%Tab. \ref{table:quality} presents the quantitative results for the overall quality of generated camouflaged images. Background-guided methods yield substantially higher FID and KID scores due to a domain gap between manually specified backgrounds and real camouflaged images. By contrast, foreground-guided methods generate better background scenes based on the objects. However, most of these methods are designed for generic image generation. Thus the distribution discrepancy still exists. In contrast, the proposed FACIG outperforms all other methods and achieves an overall improvement of 17.7$\%$ in FID and 35.5$\%$ in KID compared with the SOTA method. This suggests that FACIG is particularly effective at generating camouflaged images with higher quality. The foreground-background coherence makes the generated images more realistic than LAKE-RED.

Note that there is a performance drop for salient and general objects in foreground-guided methods. This is due to the complexity of the objects, where diverse classes and sizes make it challenging to generate suitable background scenes to seamlessly conceal them. Despite this, our method achieves optimal performance, confirming the capability to transfer images from generic to camouflaged.
\\
\noindent $\bullet$ \textbf{Foreground Fidelity.}
% Since previous work does not provide a definition, we define a small object, considering the number of objects, as one whose area is less than 1/64th of the original image, corresponding to just 2$\times$2 pixels in the latent features at the final layer of the LDM. 
In view of the lack of definitions in previous works, we define the \textbf{small object}, considering the number of objects, as one having a surface area less than 1/64th of the original image, which corresponds to just 2$\times$2 pixels in the latent features.
% The number of small objects in CO, SO, and GO subsets are 502, 261, and 510, respectively. As shown in Tab. \ref{table:foreground}, the proposed FACIG improves the foreground fidelity, especially for small objects, in all subsets. Although \cite{perceptual} has shown that the PSNR and SSIM do not adequately correlate with the human perceptual quality, our method shows consistent improvements in both metrics of the foreground objects, verifying the effectiveness of foreground reconstruction. To further demonstrate this, we present qualitative results in Fig. \ref{fig:fig1}. The image generated by LAKE-RED suffers from severe foreground distortion, while our method successfully reconstructs the foreground object.
We define a \textbf{small object} as one with a surface area less than 1/64th of the original image, equivalent to 2×2 pixels in latent features. The CO, SO, and GO subsets contain 502, 261, and 510 small objects, respectively. As shown in Tab.\ref{table:foreground}, FACIG significantly improves foreground fidelity, particularly for small objects, across all subsets. While PSNR and SSIM may not fully reflect human perceptual quality\cite{perceptual}, our method consistently improves these metrics, demonstrating effective foreground reconstruction. Qualitative results in Fig.~\ref{fig:fig1} show that LAKE-RED suffers from severe foreground distortion, while our method accurately reconstructs foreground objects.
\begin{figure*}[t] \centering
    \includegraphics[width=0.3\textwidth]{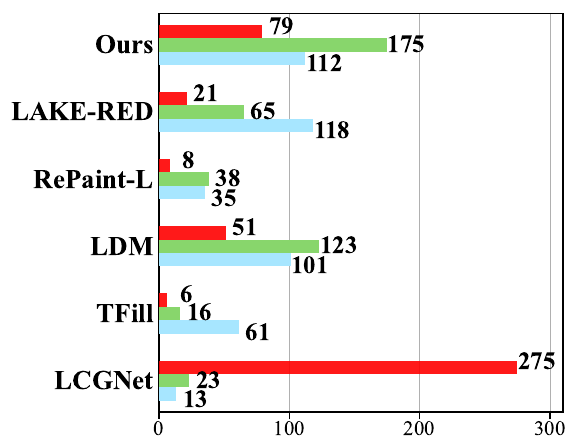}
    \includegraphics[width=0.3\textwidth]{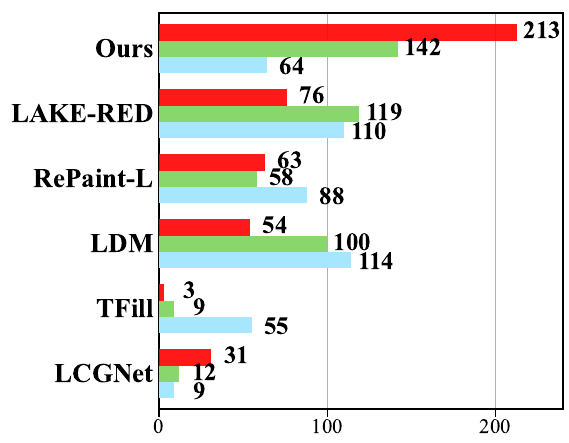}
    \includegraphics[width=0.3\textwidth]{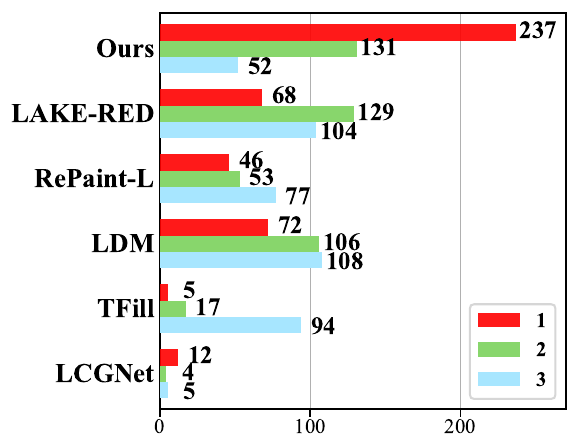}

    \makebox[0.3\textwidth][r]{\footnotesize\raisebox{0.6em}{\textbf{Q1: Which result is the hardest to find?}}}
    \makebox[0.3\textwidth][r]{\footnotesize \raisebox{-0.6em}{\textbf{\shortstack[l]{Q2: Which is the most visually natural \\ \quad \quad and reasonable result?}}}}
    \makebox[0.3\textwidth][r]{\footnotesize \raisebox{-0.6em}{\textbf{\shortstack[l]{Q3: Which result appears closest to the \\ \quad \quad real camouflaged images dataset?}}}}
    \caption {User study evaluating human preferences for camouflaged images generated by our method and SOTA methods.} 
    \label{fig:user_study}
\vspace{-0.1cm}
\end{figure*}

\noindent $\bullet$ \textbf{Visualization.} 
Fig. \ref{fig:visualization} presents qualitative comparisons across methods. LCGNet achieves strong concealment but yields unnatural results due to overly concealed foreground objects and contextual mismatches between foreground and background. Foreground-guided methods (e.g., TFill, LDM, RePaint-L) lack explicit background information, resulting in less coherent scenes. Similarly, LAKE-RED exhibits foreground-background inconsistencies. In contrast, our method effectively captures foreground features and ensures seamless integration with the background, achieving superior visual quality.

%Fig. \ref{fig:visualization} illustrates the qualitative results of various methods. LCGNet, representing background-guided methods, excels at concealing objects, the overly invisible objects appear unnatural, and contextual mismatches between backgrounds and foregrounds lead to unrealistic results. Foreground-guided methods like TFill, LDM, and RePaint-L, lacking explicit background information, often generate less authentic scenes. Similarly, LAKE-RED suffers from foreground-background inconsistency, reducing camouflage effectiveness. In contrast, our method effectively captures foreground features and generates seamlessly aligned backgrounds, producing higher-quality results.

\noindent $\bullet$ \textbf{User Study.} 
% We conduct an extensive user study to verify the overall quality of the generated camouflaged images in human perception. We randomly select 20 foreground image sets and utilize various methods to generate the results. Each participant is asked to rate the generated results based on three questions to assess the concealing capability, naturalness, and realism of the generated images.
We conducted a user study to evaluate the quality of generated camouflaged images. Twenty foreground image sets were randomly selected, and results were generated using various methods. Each participant rated the images based on concealing capability, naturalness, and realism, selecting their top three choices for each question. Feedback from 22 participants, mostly with deep learning backgrounds, was collected.
% {\footnotesize
% \begin{description}[leftmargin=!,labelwidth=\widthof{\bfseries -Q1:}]
% \setlength\itemsep{1pt}
%     \item[Q1:] \textbf{Which result is the hardest to find?}
%     \item[Q2:] \textbf{Which is the most visually natural and reasonable result?}
%     \item[Q3:] \textbf{Which result appears closest to the real camouflaged images dataset?}
% \end{description}
% }
% \noindent For each question, participants need to choose their top 3 choices, with 1 being the highest. 
% Detailed feedback was collected from 22 participants, most of whom have deep learning backgrounds. 
% As shown in Fig. \ref{fig:user_study}, LCGNet surpasses other methods in \textbf{Q1} due to the almost invisible foreground objects. However, changing foreground objects' texture makes them visually unnatural. Besides, without deliberately specified background scenes, the generated images lack contextual coherence between the foreground and background. In contrast, our generated images are more visually natural than those from other methods. With our FAFIM module, the generated background scenes reach higher visual consistency with the foreground objects, leading to more realistic results.
As shown in Fig.~\ref{fig:user_study}, LCGNet excels in \textbf{Q1} for concealing foreground objects but produces visually unnatural textures and lacks contextual coherence. In contrast, our method generates more natural and realistic images, with the FAFIM module ensuring higher foreground-background consistency.

% \subsection{Ablation Study of Component Effectiveness}
% We gradually add each component of our proposed method to verify its effectiveness. As shown in Tab. \ref{table:component}, $\mathcal{L}_{FADL}$ significantly boosts foreground fidelity, thereby improving overall image quality. Adding the FAFIM further elevates image quality. Finally, incorporating positional encoding into FAFIM, which captures the locality relationship between the background and foreground, achieves the best results. Other ablation studies are provided in the supplementary.
\subsection{Ablation Study of Component Effectiveness}  
We incrementally add components of our method to validate their effectiveness. As shown in Tab.~\ref{table:component}, $\mathcal{L}_{FADL}$ significantly enhances foreground fidelity, improving overall image quality. Adding FAFIM further boosts quality, and incorporating positional encoding into FAFIM, which captures background-foreground locality, achieves the best results. Additional ablation studies are in the supplementary.
% We present the overall performance across all subsets.
\begin{table}[h]
    \small
    \centering
    \tabcolsep=2.2pt
    \renewcommand\arraystretch{0.9}
    \caption{Ablation study of different components. \enquote{$PE$} refers to positional encoding in FAFIM.}
    \begin{tabular}{ccc|cccc}
    \toprule[1pt]
    $\mathcal{L}_{FADL}$ &FAFIM &$PE$ &FID $\downarrow$ & KID $\downarrow$ &PSNR (f)$\uparrow$ &SSIM (f)$\uparrow$ \\ 
    \midrule[1pt]
    \multicolumn{3}{c|}{Baseline} &64.27 &0.0355 &20.03 &0.795\\	
    \checkmark &- &- &58.10	&0.0266	&23.19	&0.881\\
    \checkmark &\checkmark &- &54.03 &0.0239 &23.18	&0.880\\
    \checkmark &\checkmark &\checkmark &\textbf{52.87} &\textbf{0.0229} &\textbf{23.45} &\textbf{0.886} \\
    \midrule[1pt] 
    \end{tabular}
\label{table:component}
%\vspace{-0.2cm}
\end{table}

\section{Conclusion}
% In this paper, we propose a foreground-aware camouflaged image generation (FACIG) model by using latent diffusion. We introduce FAFIM to effectively enhance the integration between foreground features and background knowledge, mitigating the lack of foreground-background coherence. Additionally, we design a foreground-aware denoising loss to emphasize foreground reconstruction, reducing foreground distortion. Our proposed FACIG model outperforms state-of-the-art methods, generating camouflaged images with superior quality and foreground fidelity.
In this paper, we propose the FACIG model for camouflaged image generation using latent diffusion. Our FAFIM module enhances foreground-background integration, addressing coherence issues, while a foreground-aware denoising loss improves foreground reconstruction and reduces distortion. FACIG surpasses state-of-the-art methods, delivering high-quality camouflaged images with improved foreground fidelity.

\bibliographystyle{IEEEbib}
\bibliography{main}

\end{document}

\title{Foreground Focus: Enhancing Coherence and Fidelity in Camouflaged Image Generation}

\author{Supplementary Material}

\maketitle
\renewcommand\thesection{\Alph{section}}
This supplementary material includes: 1) details about the experimental settings, 2) more ablation studies about the design of the loss function and the hyperparameters of our method, 3) more qualitative comparison of our method and LAKE-RED, and 4) details about the user study.
\section{Experimental Settings}
$\bullet$ \textbf{Implementation Details.}
We implement our model using PyTorch on NVIDIA Tesla V100 GPUs and A100 GPUs. The model is initialized with a pre-trained LDM and incorporates a pre-trained VQ-VAE as the autoencoder. During training, the autoencoder and codebook remain frozen. The model undergoes 50K iterations with a batch size of 8, and other hyperparameters are set similarly to those in LAKE-RED.

\noindent $\bullet$ \textbf{Datasets} Following LAKE-RED~\cite{lake}, we utilize 4040 real images (3040 from COD10K~\cite{COD10K} and 1000 from CAMO~\cite{CAMO}) for training our model. In evaluation, we collect three subsets of data as testing data: Camouflaged Objects (CO), Salient Objects (SO), and General Objects (GO). The CO subset comprises 6473 images from the COD10K, CAMO, and NC4K datasets~\cite{NC4K}. For the SO and GO subsets, we employ the datasets collected by LAKE-RED, which include randomly selected images from DUTS~\cite{duts}, DUT-OMRON~\cite{dut-omron}, and COCO2017~\cite{coco}, among others. SO and GO subsets are used to evaluate the model's capability for transferring images from non-camouflaged to camouflaged.

\section{Ablation study}
$\bullet$ \textbf{Analysis of loss function.}
We compared three alternative function types for designing $\omega$ with our method, with the function curves depicted in Fig. \ref{fig:loss}. The \enquote{Log} function is defined as $1-2log(r)$, and the \enquote{Reciprocal} function as $1/r$. The results are shown in Tab. \ref{table:loss}. Without the upper bound $\alpha$, the \enquote{Reciprocal} function results in an excessively large $\omega$ for small objects, destabilizing the training and leading to the poorest performance. While the \enquote{Linear} function still suffers from overweighting, the \enquote{Log} function provides insufficient upweighting. In contrast, our method strikes an optimal balance, achieving the best overall performance.
% \vspace{-5pt}

\noindent $\bullet$ \textbf{Selection of upper bound for $\omega$.}  As shown in Tab. \ref{table:alpha}, we try different choices of $\alpha$ in designing weighting factor $\omega$ within $\mathcal{L}_{FADL}$. The reciprocal of $\alpha$ determines the upper bound of $\omega$. We find $\alpha=1/8$ works best. Note that for $\alpha=1/16$, the excessively large $\omega$ for small objects destabilizes the training process, leading to performance drop.
\\ \noindent $\bullet$ \textbf{Selection of patch size for FAFIM.}  We try different choices of patch size in FAFIM. Based on the result illustrated in Tab. \ref{table:patch_size}, we choose $P=4$ as the final patch size in our method.
\setcounter{figure}{4}
\begin{figure}[h]
    \centering
    \includegraphics[width=0.38\textwidth]{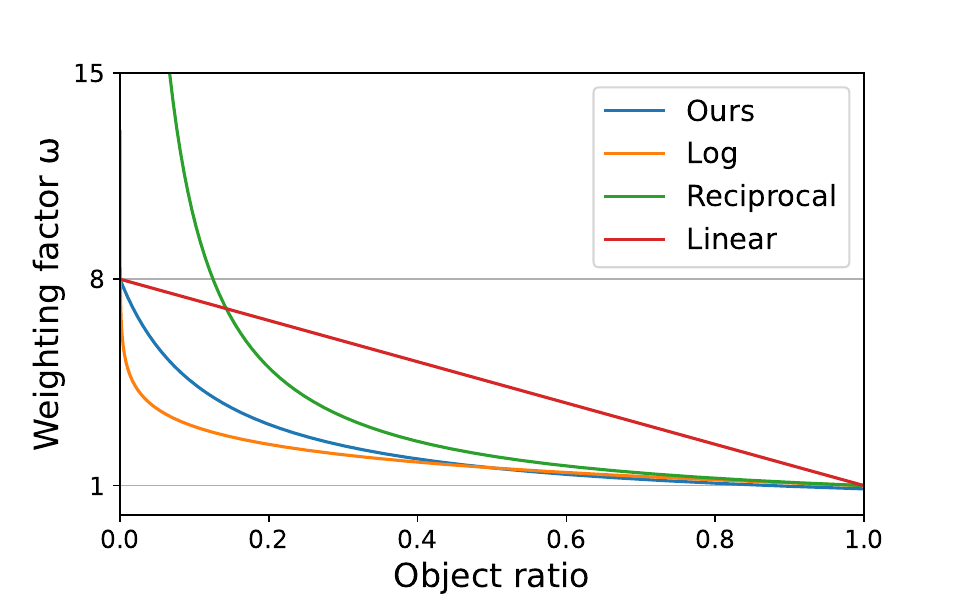}
    \caption{Plot of various functions in the design of $\omega$, showing respective weighting factors as a function of object ratio.}
    \label{fig:loss}
\end{figure}

\setcounter{table}{3}
\begin{table}[h]
    \centering
    \small
    \tabcolsep=4pt
    \renewcommand\arraystretch{0.9}
    \caption{Ablation study of different functions in the design of $\omega$. We present the overall performance across all subsets.}
    \begin{tabular}{c|cccc}
    \toprule[1pt]
    Function&FID $\downarrow$ & KID $\downarrow$ &PSNR (f)$\uparrow$ &SSIM (f)$\uparrow$ \\ 
    \midrule[1pt]
    Linear &55.67 &0.0248 &22.84 &0.874\\
    Log &53.99	&0.0237 &22.76 &0.872\\
    Reciprocal &63.25 &0.0318 &22.79 &0.872\\
    \textbf{Ours} &\textbf{52.87} &\textbf{0.0229} &\textbf{23.45} &\textbf{0.886} \\
    \midrule[1pt] 
    \end{tabular}
\label{table:loss}
\end{table}
\begin{table}[h]
    \centering
    \tabcolsep=4pt
    \renewcommand\arraystretch{1.0}
    \caption{Ablation study on $\alpha$ selection for $\mathcal{L}_{FADL}$.}
    \begin{tabular}{c|cccc}
    \toprule[1pt]
    $\alpha$ &FID $\downarrow$ & KID $\downarrow$ &PSNR (f)$\uparrow$  &SSIM (f)$\uparrow$ \\ 
    \midrule[1pt]
    1/4 &55.25 &0.0249 &23.34  &0.883\\
    \textbf{1/8} &\textbf{52.87} &\textbf{0.0229} &\textbf{23.45} &\textbf{0.886} \\
    1/16 &55.71	&0.0252 &22.99	&0.878\\
    \midrule[1pt] 
    \end{tabular}
\label{table:alpha}
\end{table}
\begin{table}[h]
    \centering
    \tabcolsep=4pt
    \renewcommand\arraystretch{1.0}
    \caption{Ablation study on the patch size $P$ for FAFIM.}
    \begin{tabular}{c|cccc}
    \toprule[1pt]
    $P$ &FID $\downarrow$ & KID $\downarrow$ &PSNR (f)$\uparrow$  &SSIM (f)$\uparrow$ \\ 
    \midrule[1pt]
    2 &61.56 &0.0300 &22.91	&0.875\\
    \textbf{4} &\textbf{52.87} &\textbf{0.0229} &\textbf{23.45} &\textbf{0.886} \\
    8 &55.92 &0.0255 &23.21	&0.882\\
    \midrule[1pt] 
    \end{tabular}
\label{table:patch_size}
\end{table}
\section{More Visualization}
\begin{figure*}[b]
    \centering
    \includegraphics[width=0.15\textwidth,height=0.15\textwidth]{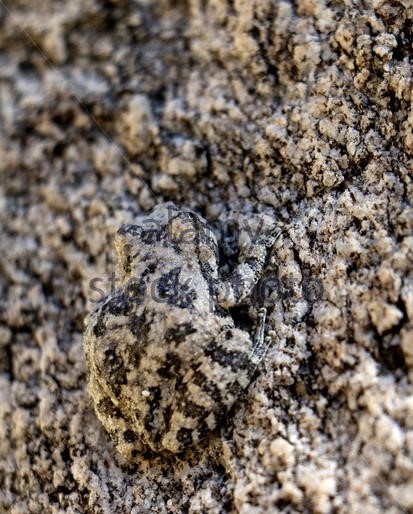}
    \includegraphics[width=0.15\textwidth,height=0.15\textwidth]{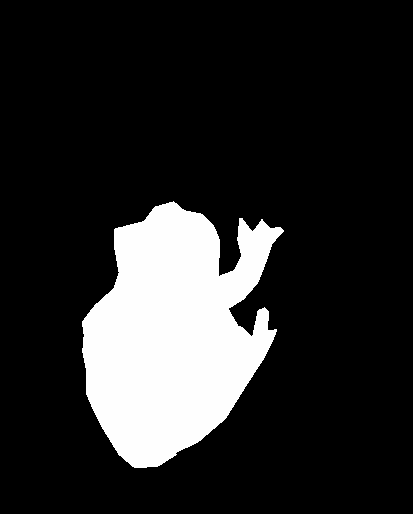}
    \includegraphics[width=0.15\textwidth,height=0.15\textwidth]{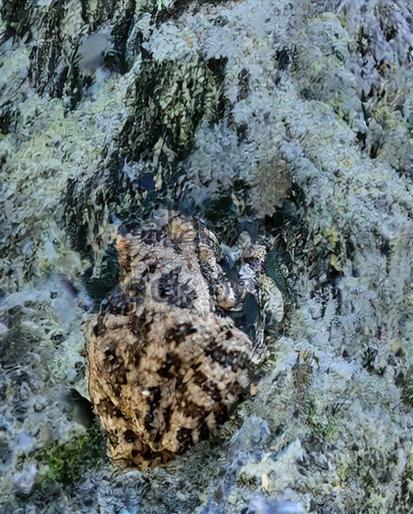}
    \includegraphics[width=0.15\textwidth,height=0.15\textwidth]{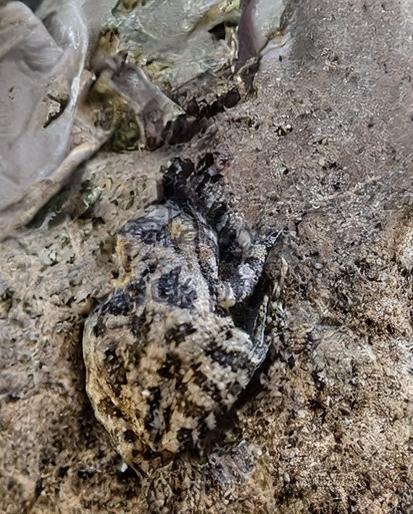}
    \\
    \includegraphics[width=0.15\textwidth,height=0.15\textwidth]{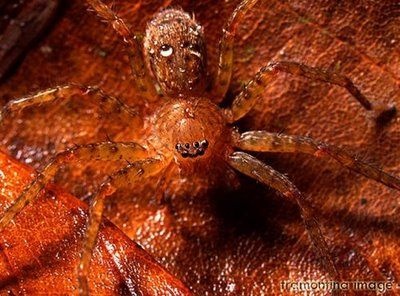}
    \includegraphics[width=0.15\textwidth,height=0.15\textwidth]{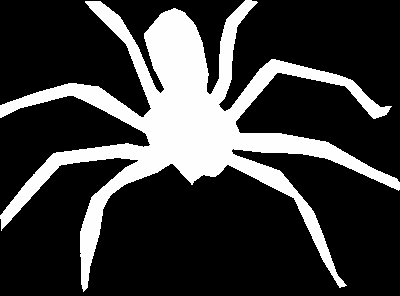}
    \includegraphics[width=0.15\textwidth,height=0.15\textwidth]{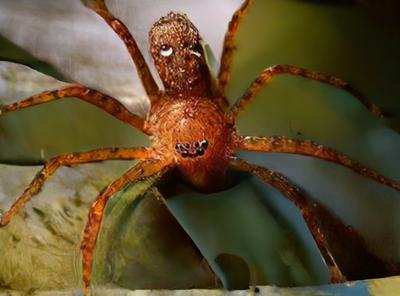}
    \includegraphics[width=0.15\textwidth,height=0.15\textwidth]{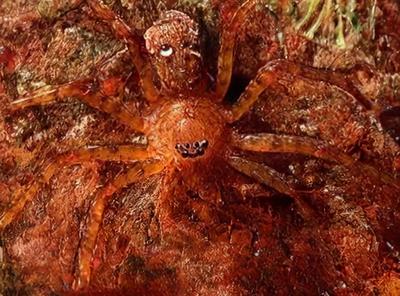}
    \\
    \includegraphics[width=0.15\textwidth,height=0.15\textwidth]{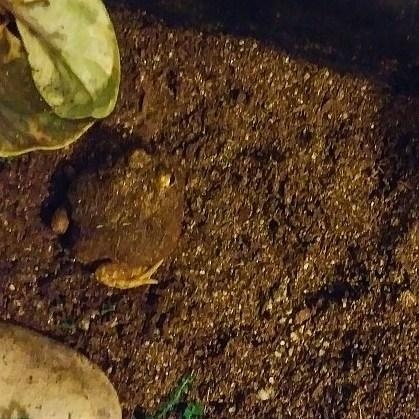}
    \includegraphics[width=0.15\textwidth,height=0.15\textwidth]{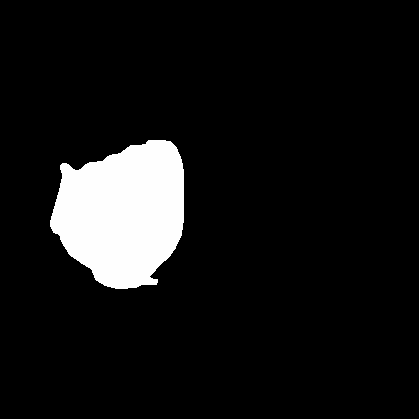}
    \includegraphics[width=0.15\textwidth,height=0.15\textwidth]{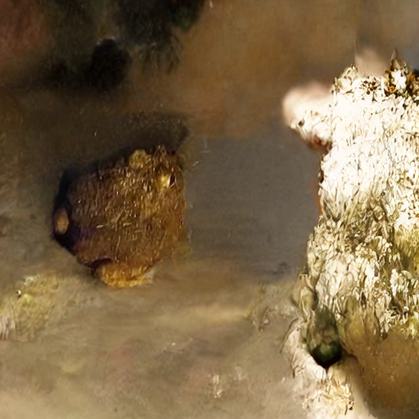}
    \includegraphics[width=0.15\textwidth,height=0.15\textwidth]{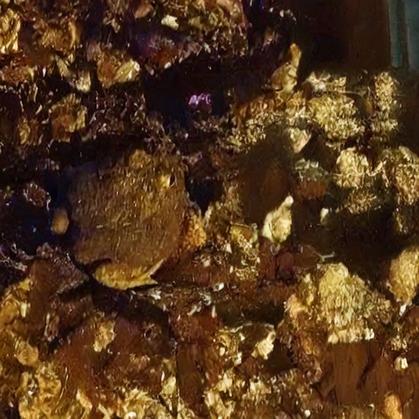}
    \\
    \includegraphics[width=0.15\textwidth,height=0.15\textwidth]{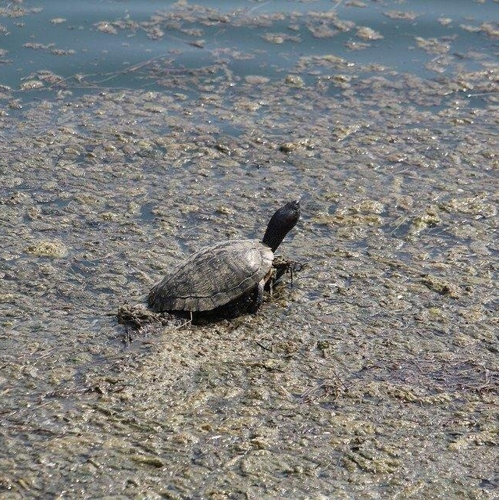}
    \includegraphics[width=0.15\textwidth,height=0.15\textwidth]{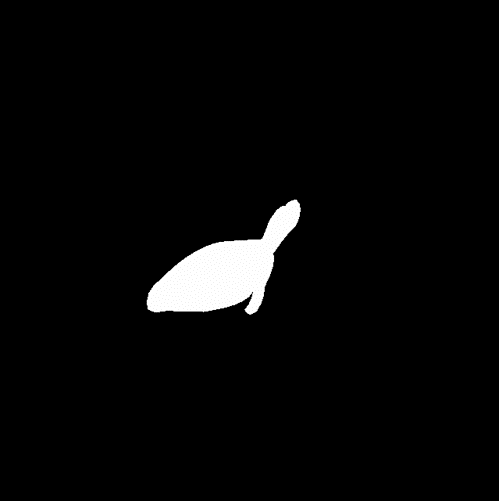}
    \includegraphics[width=0.15\textwidth,height=0.15\textwidth]{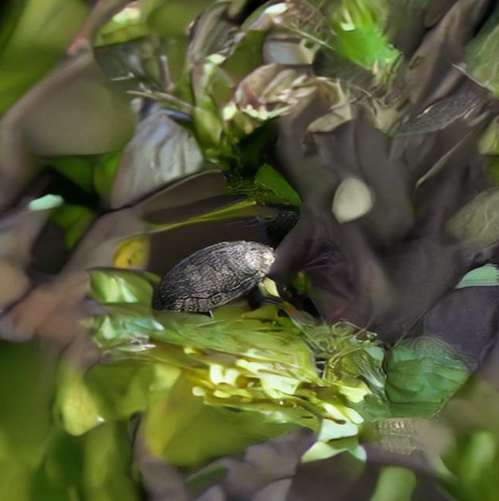}
    \includegraphics[width=0.15\textwidth,height=0.15\textwidth]{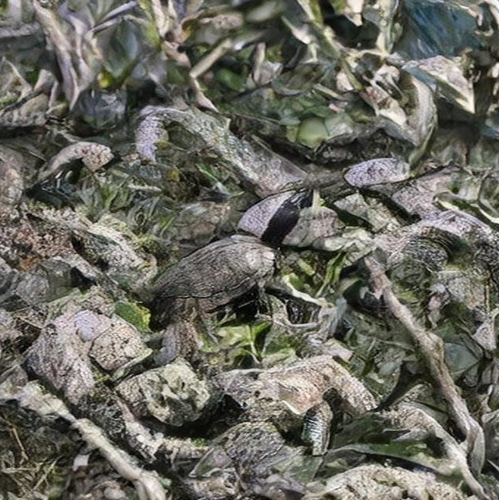}
    \\
    \includegraphics[width=0.15\textwidth,height=0.15\textwidth]{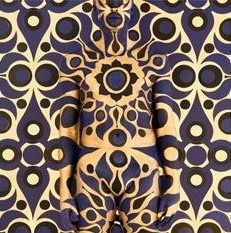}
    \includegraphics[width=0.15\textwidth,height=0.15\textwidth]{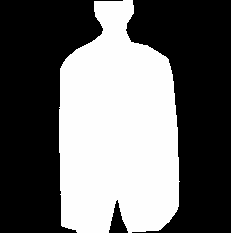}
    \includegraphics[width=0.15\textwidth,height=0.15\textwidth]{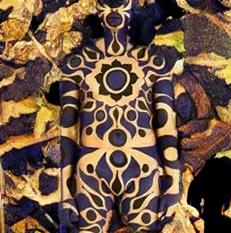}
    \includegraphics[width=0.15\textwidth,height=0.15\textwidth]{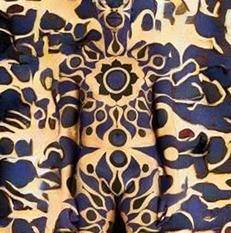}
    \\
    \includegraphics[width=0.15\textwidth,height=0.15\textwidth]{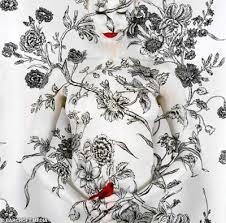}
    \includegraphics[width=0.15\textwidth,height=0.15\textwidth]{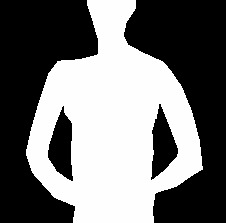}
    \includegraphics[width=0.15\textwidth,height=0.15\textwidth]{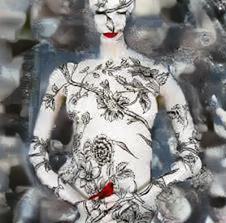}
    \includegraphics[width=0.15\textwidth,height=0.15\textwidth]{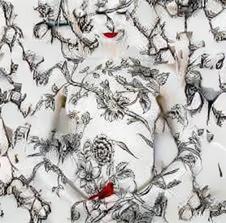}
    \\

    \vspace{5pt}
    \makebox[0.15\textwidth]{\Large Image}
    \makebox[0.15\textwidth]{\Large Mask}
    \makebox[0.15\textwidth]{\Large LAKE-RED}
    \makebox[0.15\textwidth]{\Large Ours}
    \caption{Qualitative results of our method and LAKE-RED in Camouflaged Objects subset.}
    \label{fig:COD_vis}
\end{figure*}

\begin{figure*}[t]
    \centering
    \includegraphics[width=0.15\textwidth,height=0.15\textwidth]{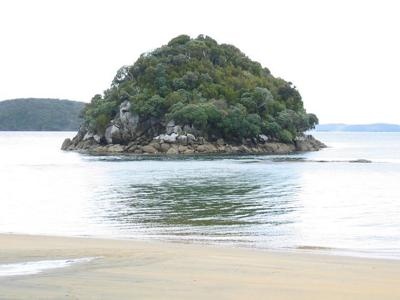}
    \includegraphics[width=0.15\textwidth,height=0.15\textwidth]{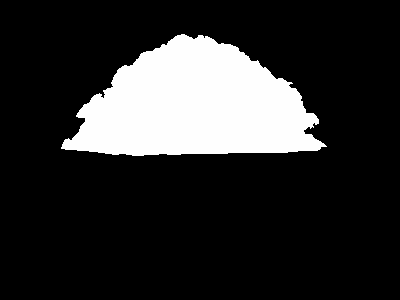}
    \includegraphics[width=0.15\textwidth,height=0.15\textwidth]{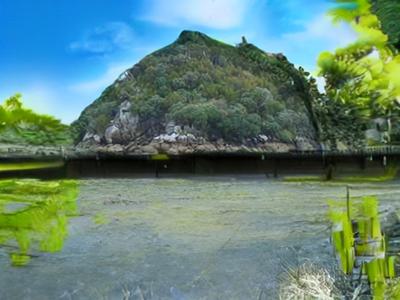}
    \includegraphics[width=0.15\textwidth,height=0.15\textwidth]{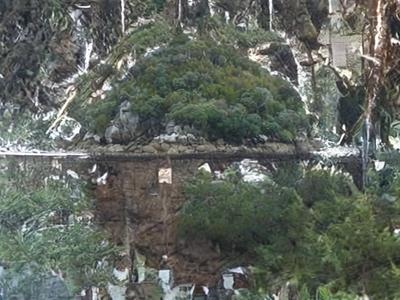}
    \\
    \includegraphics[width=0.15\textwidth,height=0.15\textwidth]{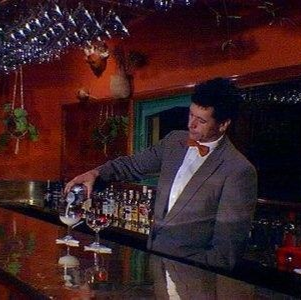}
    \includegraphics[width=0.15\textwidth,height=0.15\textwidth]{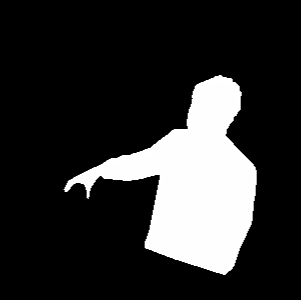}
    \includegraphics[width=0.15\textwidth,height=0.15\textwidth]{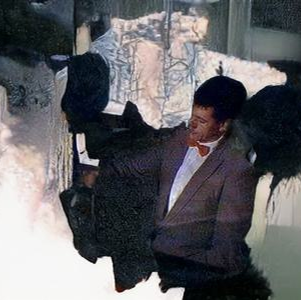}
    \includegraphics[width=0.15\textwidth,height=0.15\textwidth]{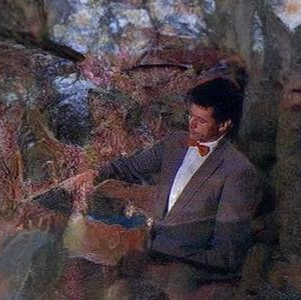}
    \\
    \includegraphics[width=0.15\textwidth,height=0.15\textwidth]{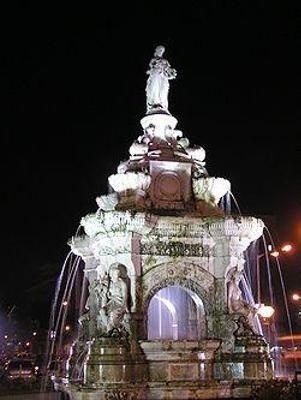}
    \includegraphics[width=0.15\textwidth,height=0.15\textwidth]{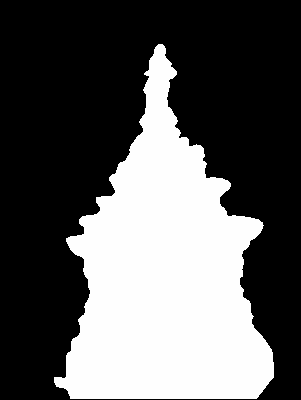}
    \includegraphics[width=0.15\textwidth,height=0.15\textwidth]{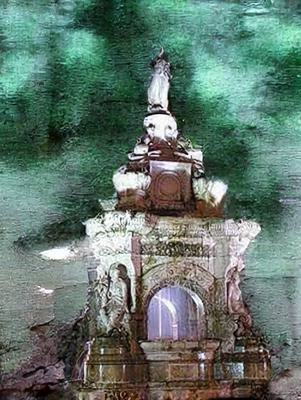}
    \includegraphics[width=0.15\textwidth,height=0.15\textwidth]{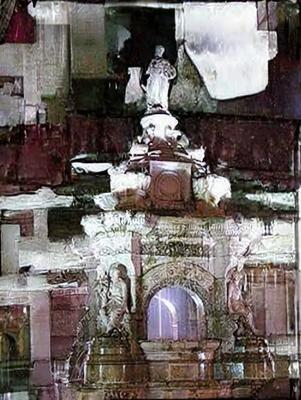}
    \\
    \includegraphics[width=0.15\textwidth,height=0.15\textwidth]{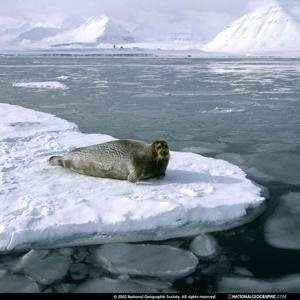}
    \includegraphics[width=0.15\textwidth,height=0.15\textwidth]{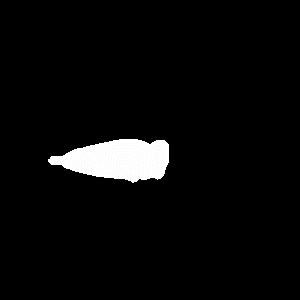}
    \includegraphics[width=0.15\textwidth,height=0.15\textwidth]{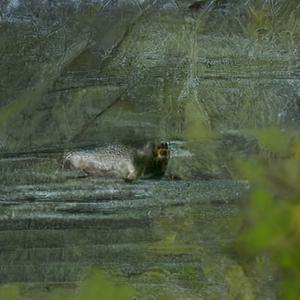}
    \includegraphics[width=0.15\textwidth,height=0.15\textwidth]{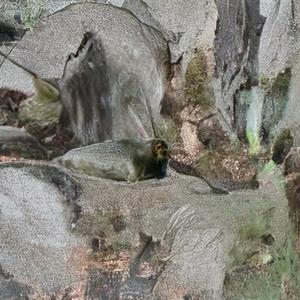}
    \\
    \includegraphics[width=0.15\textwidth,height=0.15\textwidth]{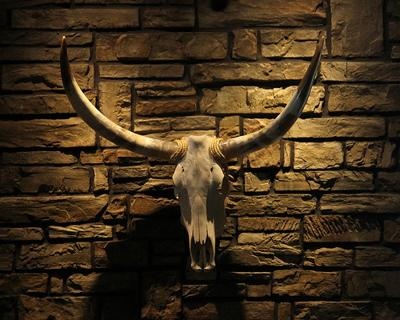}
    \includegraphics[width=0.15\textwidth,height=0.15\textwidth]{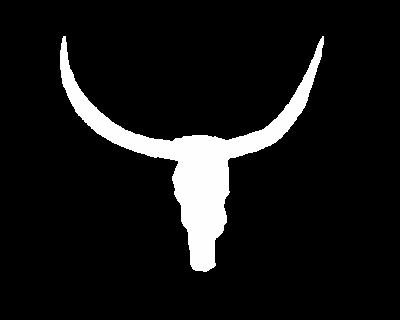}
    \includegraphics[width=0.15\textwidth,height=0.15\textwidth]{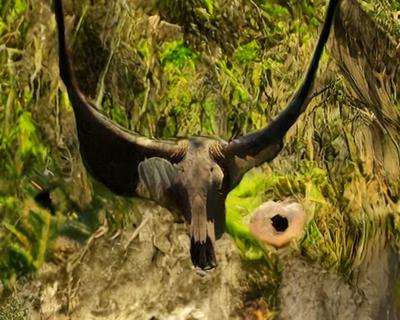}
    \includegraphics[width=0.15\textwidth,height=0.15\textwidth]{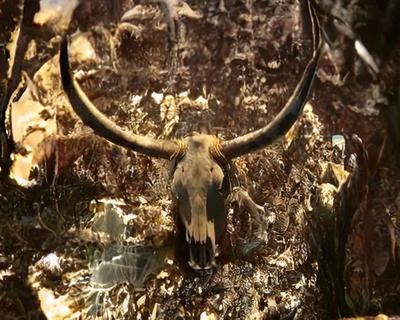}
    \\
    \includegraphics[width=0.15\textwidth,height=0.15\textwidth]{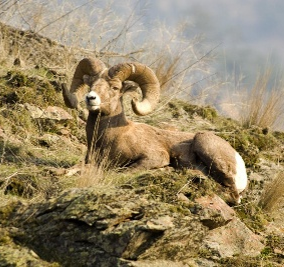}
    \includegraphics[width=0.15\textwidth,height=0.15\textwidth]{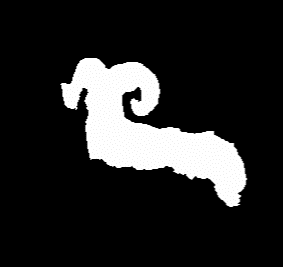}
    \includegraphics[width=0.15\textwidth,height=0.15\textwidth]{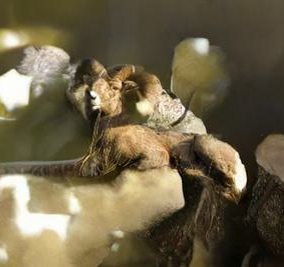}
    \includegraphics[width=0.15\textwidth,height=0.15\textwidth]{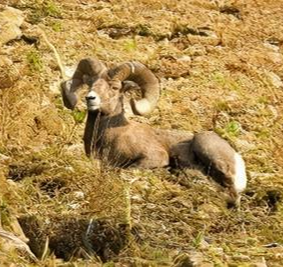}
    \\

    \vspace{5pt}
    \makebox[0.15\textwidth]{\Large Image}
    \makebox[0.15\textwidth]{\Large Mask}
    \makebox[0.15\textwidth]{\Large LAKE-RED}
    \makebox[0.15\textwidth]{\Large Ours}
    \caption{Qualitative results of our method and LAKE-RED in Salient Objects subset.}
    \label{fig:SOD_vis}
\end{figure*}

\begin{figure*}[t]
    \centering
    \includegraphics[width=0.15\textwidth,height=0.15\textwidth]{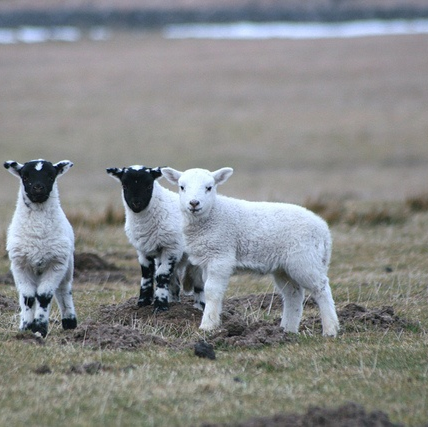}
    \includegraphics[width=0.15\textwidth,height=0.15\textwidth]{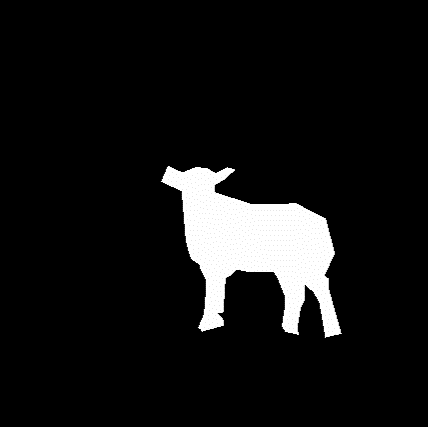}
    \includegraphics[width=0.15\textwidth,height=0.15\textwidth]{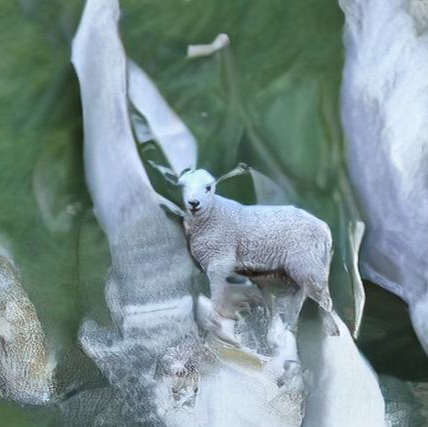}
    \includegraphics[width=0.15\textwidth,height=0.15\textwidth]{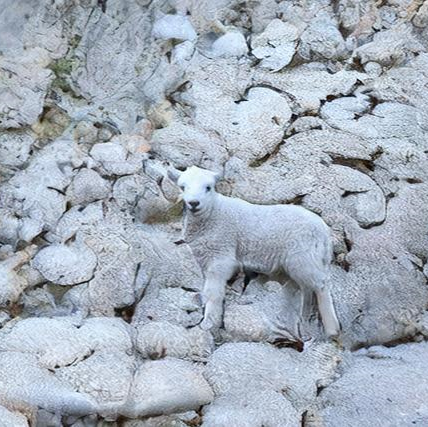}
    \\
    \includegraphics[width=0.15\textwidth,height=0.15\textwidth]{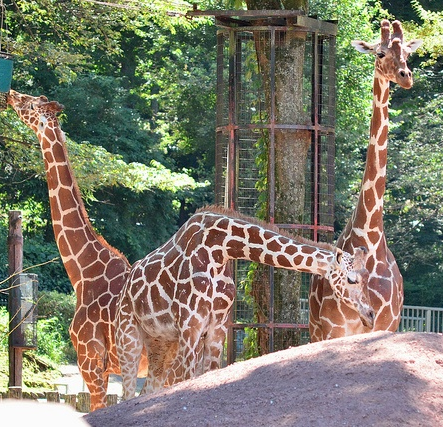}
    \includegraphics[width=0.15\textwidth,height=0.15\textwidth]{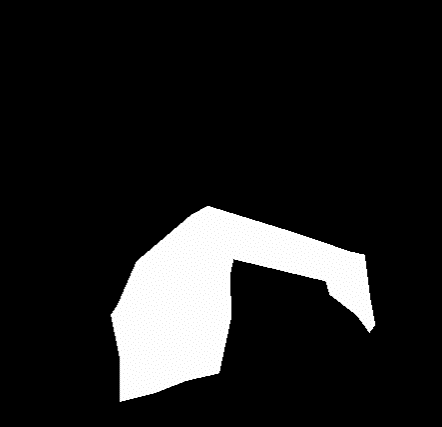}
    \includegraphics[width=0.15\textwidth,height=0.15\textwidth]{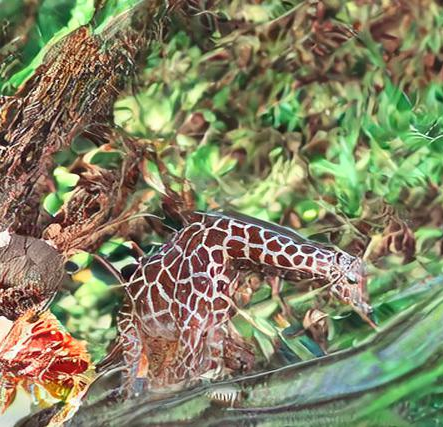}
    \includegraphics[width=0.15\textwidth,height=0.15\textwidth]{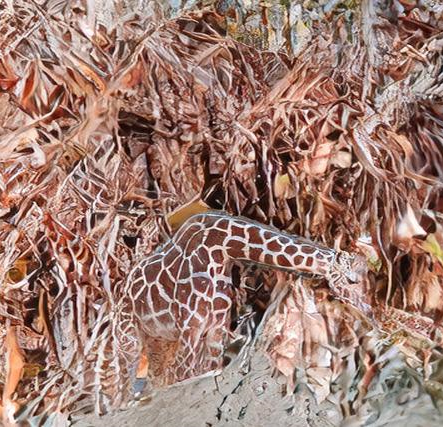}
    \\
    \includegraphics[width=0.15\textwidth,height=0.15\textwidth]{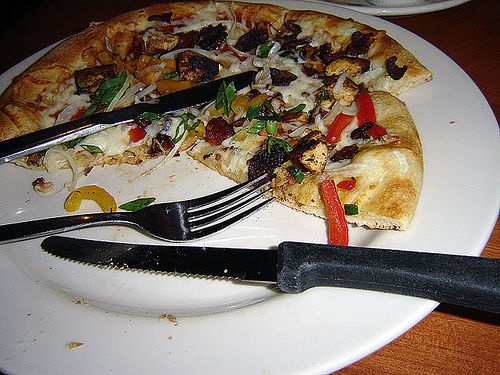}
    \includegraphics[width=0.15\textwidth,height=0.15\textwidth]{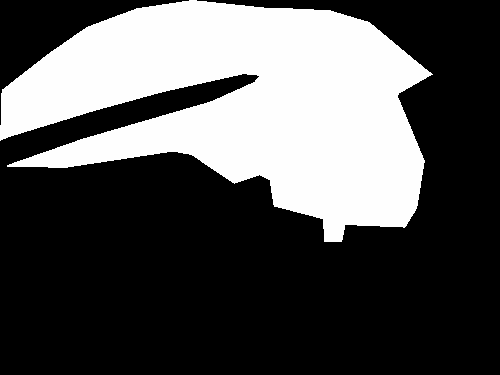}
    \includegraphics[width=0.15\textwidth,height=0.15\textwidth]{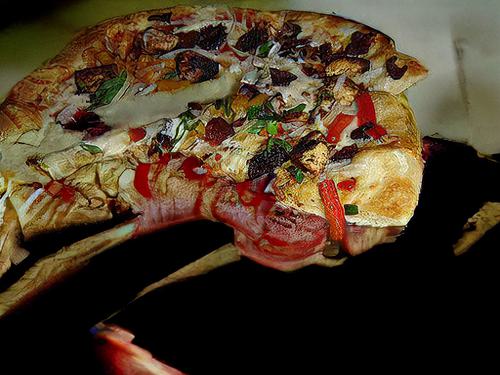}
    \includegraphics[width=0.15\textwidth,height=0.15\textwidth]{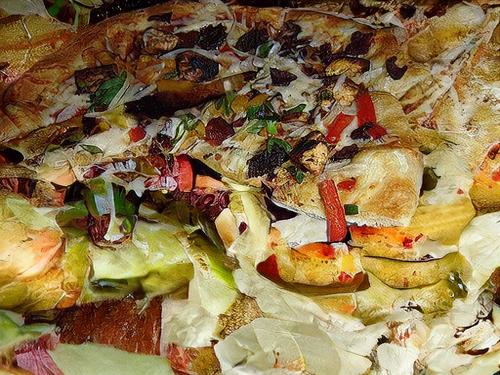}
    \\
    \includegraphics[width=0.15\textwidth,height=0.15\textwidth]{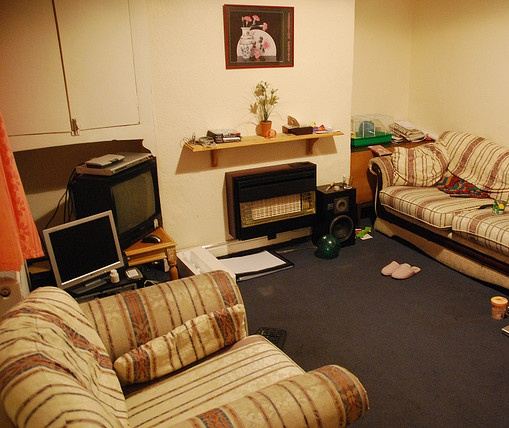}
    \includegraphics[width=0.15\textwidth,height=0.15\textwidth]{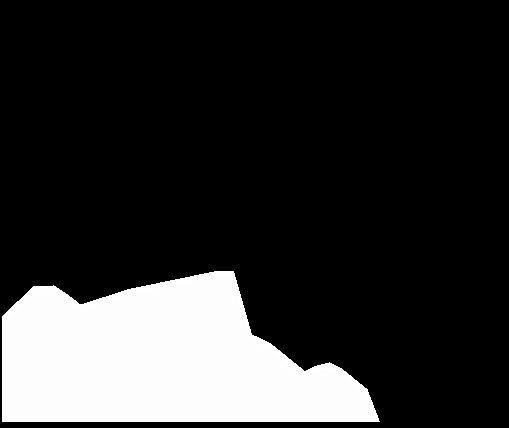}
    \includegraphics[width=0.15\textwidth,height=0.15\textwidth]{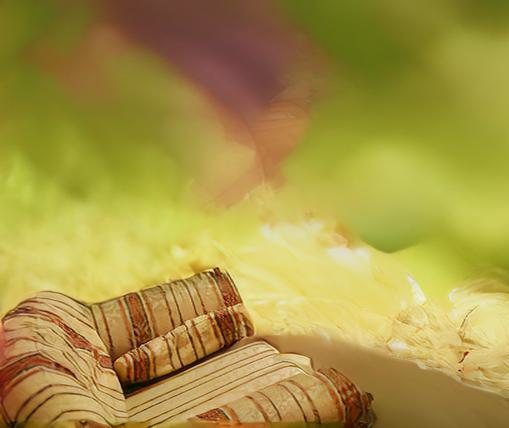}
    \includegraphics[width=0.15\textwidth,height=0.15\textwidth]{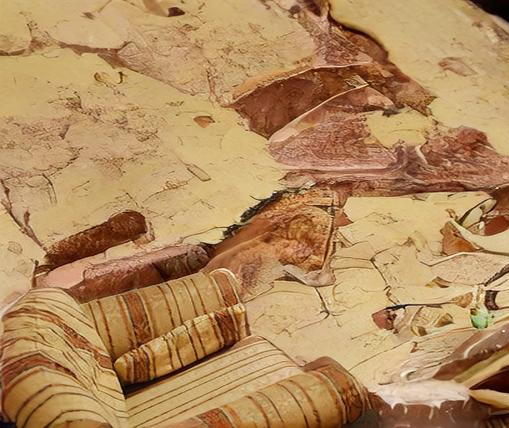}
    \\
    \includegraphics[width=0.15\textwidth,height=0.15\textwidth]{Supp_Figures/SEG/SEG_6/original.png}
    \includegraphics[width=0.15\textwidth,height=0.15\textwidth]{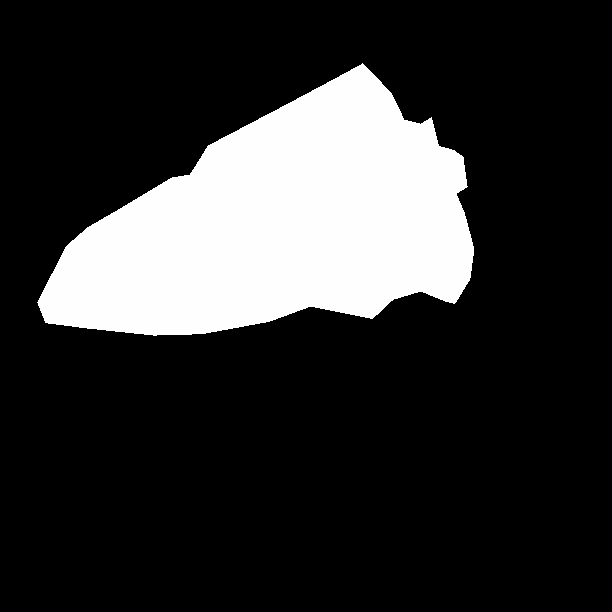}
    \includegraphics[width=0.15\textwidth,height=0.15\textwidth]{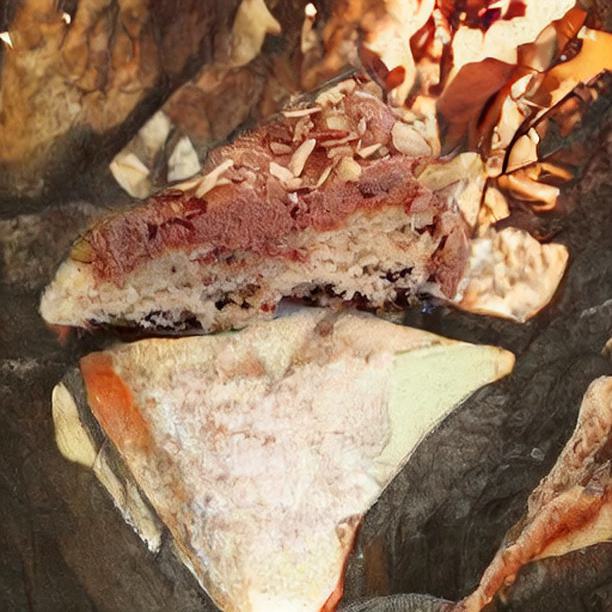}
    \includegraphics[width=0.15\textwidth,height=0.15\textwidth]{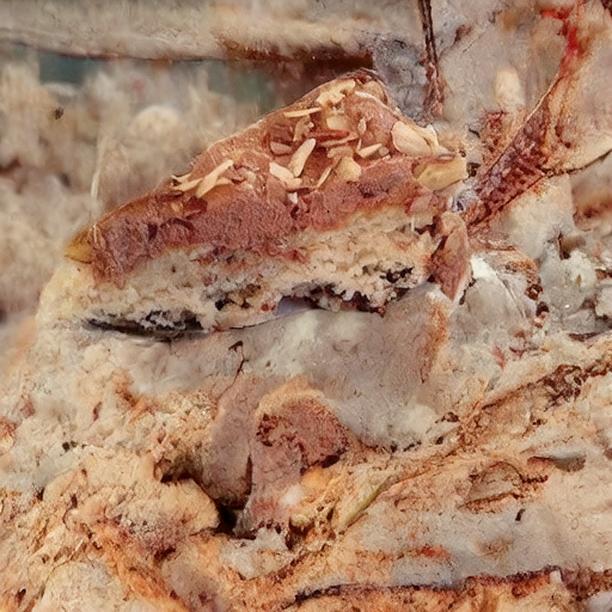}
    \\
    \includegraphics[width=0.15\textwidth,height=0.15\textwidth]{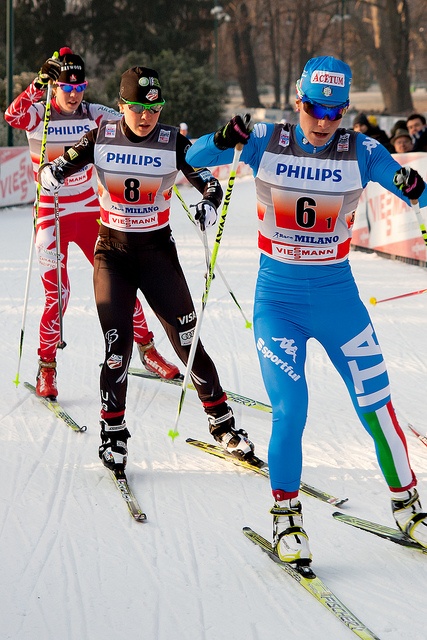}
    \includegraphics[width=0.15\textwidth,height=0.15\textwidth]{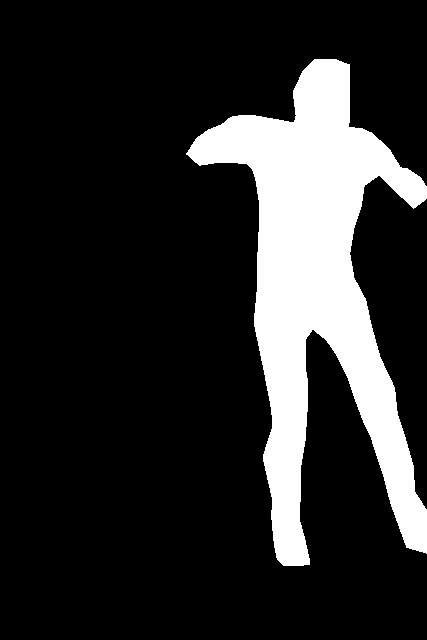}
    \includegraphics[width=0.15\textwidth,height=0.15\textwidth]{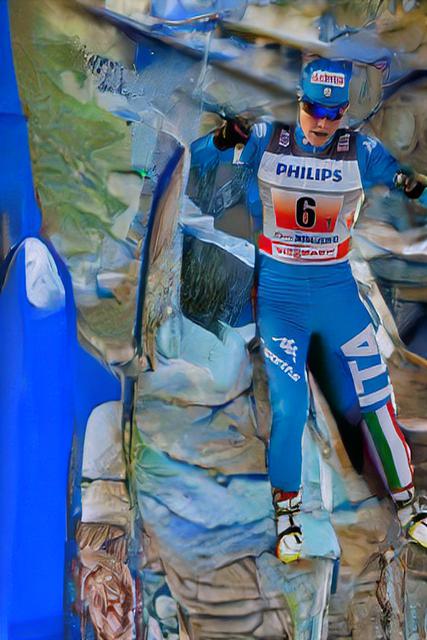}
    \includegraphics[width=0.15\textwidth,height=0.15\textwidth]{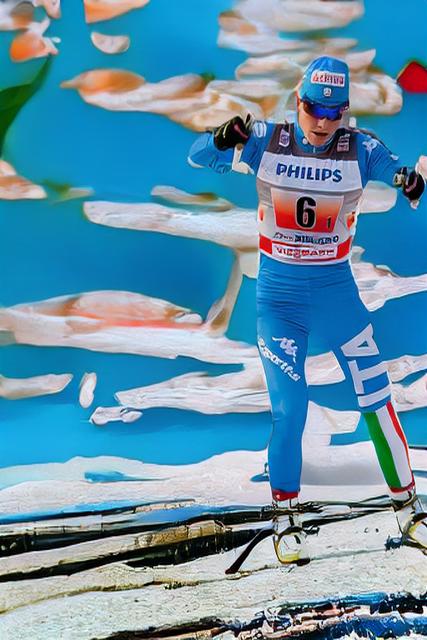}
    \\

    \vspace{5pt}
    \makebox[0.15\textwidth]{\Large Image}
    \makebox[0.15\textwidth]{\Large Mask}
    \makebox[0.15\textwidth]{\Large LAKE-RED}
    \makebox[0.15\textwidth]{\Large Ours}
    \caption{Qualitative results of our method and LAKE-RED in General Objects subset.}
    \label{fig:SEG_vis}
\end{figure*}

In this section, we provide more qualitative comparisons of our method and LAKE-RED. 
\\ \noindent $\bullet$ \textbf{Comparison with objects of varying sizes.}
Fig. \ref{fig:object_size} illustrates the comparison for objects of varying sizes.  For large objects, LAKE-RED reconstructs the foreground with visually imperceptible distortion. However, for small objects, the limited supervision in the foreground leads to severe distortion. In contrast, our foreground-aware denoising loss improves foreground reconstruction, particularly for small objects, achieving higher fidelity. To further highlight this improvement, we present more comparisons focusing on foreground fidelity for small objects in Fig. \ref{fig:fidelity}. While LAKE-RED suffers from substantial foreground distortion, our method effectively preserves foreground details with high fidelity. 
\\ \noindent $\bullet$ \textbf{Comparison of overall camouflaged image quality.}
Fig. \ref{fig:COD_vis}, Fig. \ref{fig:SOD_vis}, and Fig. \ref{fig:SEG_vis} illustrate the comparison of overall camouflaged image quality in CO, SO, and GO subsets, respectively. As shown in the three figures, the images generated by LAKE-RED lack foreground-background coherence. For example, the tone or the texture of the foreground is different from that of the background. This phenomenon is more noticeable in the last two samples in Fig. \ref{fig:COD_vis}, where LAKE-RED fails to extend the intricate foreground texture to the background scene, resulting in a disjointed appearance. In contrast, our method effectively integrates and extends the foreground texture into the generated background, achieving a more coherent and realistic result.
\begin{figure}[h!]
    \centering
    \includegraphics[width=0.42\textwidth]{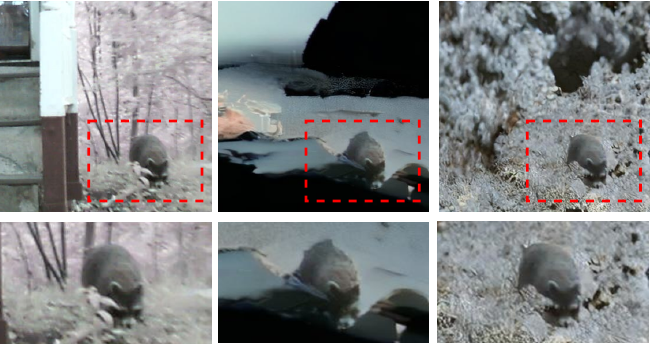}\\
    \vspace{5pt}
    \includegraphics[width=0.42\textwidth]{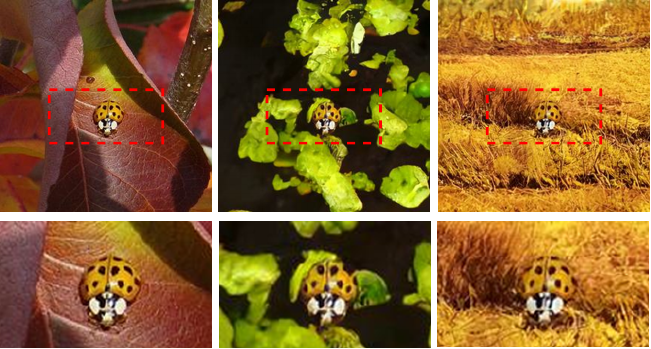}\\
    \makebox[0.14\textwidth]{Original image}
    \makebox[0.14\textwidth]{LAKE-RED}
    \makebox[0.14\textwidth]{\textbf{Ours}}
    \caption{Comparison of our method with LAKE-RED in small object reconstruction. For each image set, the top row presents the complete images, while the bottom row presents zoomed-in views of the highlighted areas.}
    \label{fig:fidelity}
\end{figure}
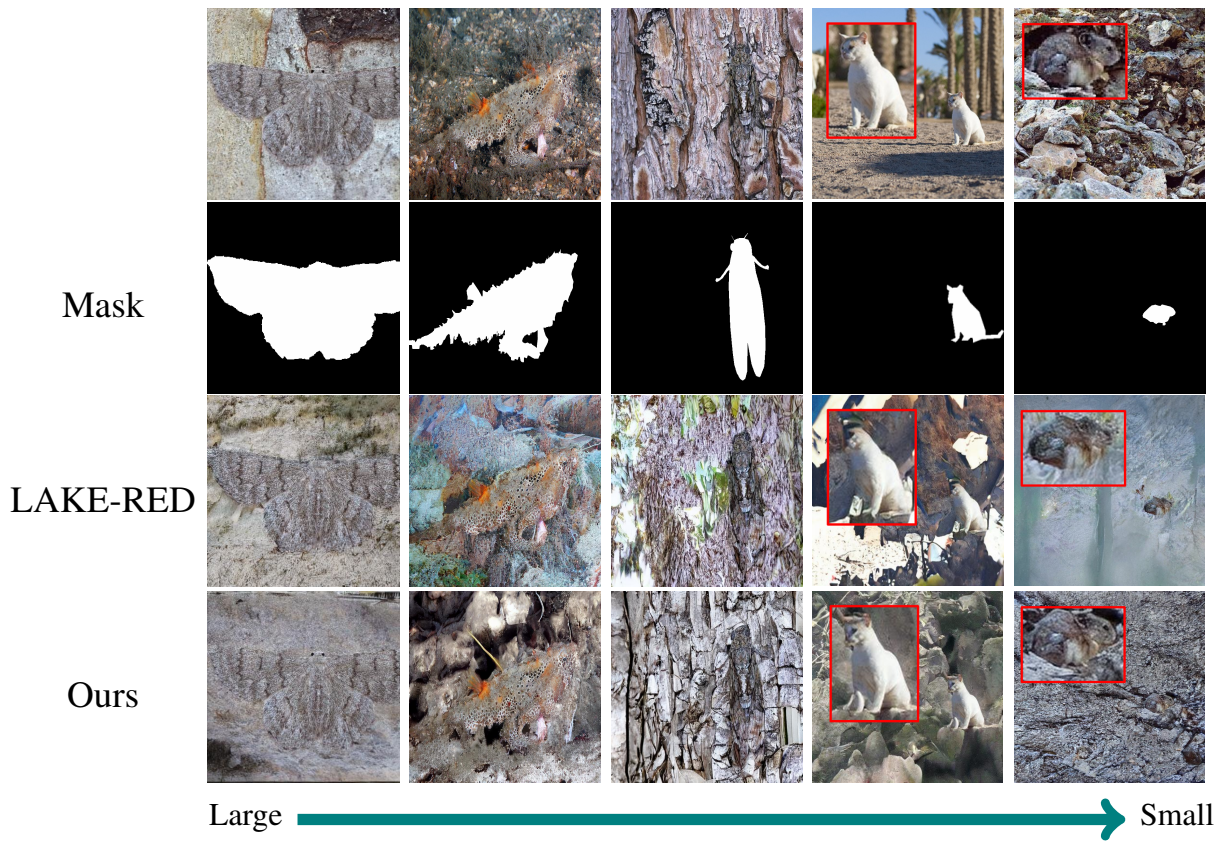
\begin{figure*}[t]
    \centering
    \raisebox{1\height}{\makebox[0.14\textwidth][]{\shortstack[c]{\Large Original \\ \Large image}}}
    \includegraphics[width=0.14\textwidth,height=0.14\textwidth]{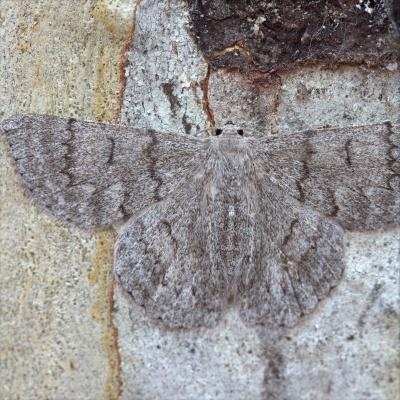}
    \hspace{-3pt} \includegraphics[width=0.14\textwidth,height=0.14\textwidth]{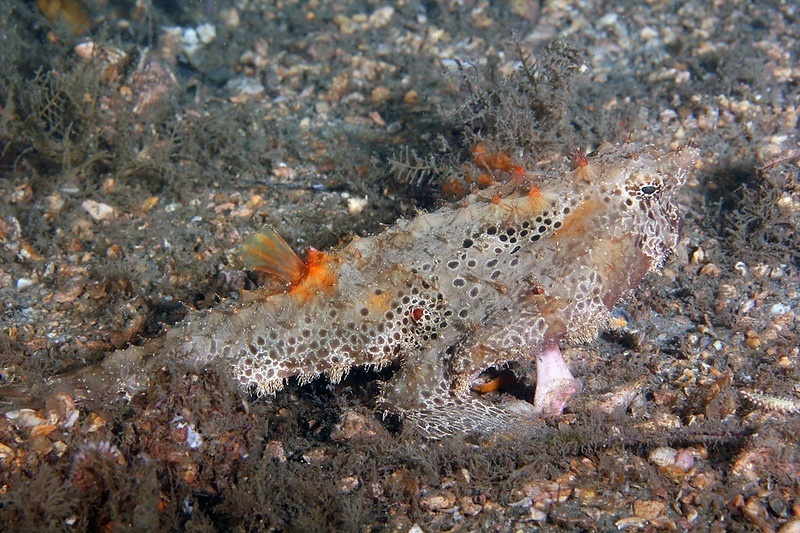}
    \hspace{-3pt} \includegraphics[width=0.14\textwidth,height=0.14\textwidth]{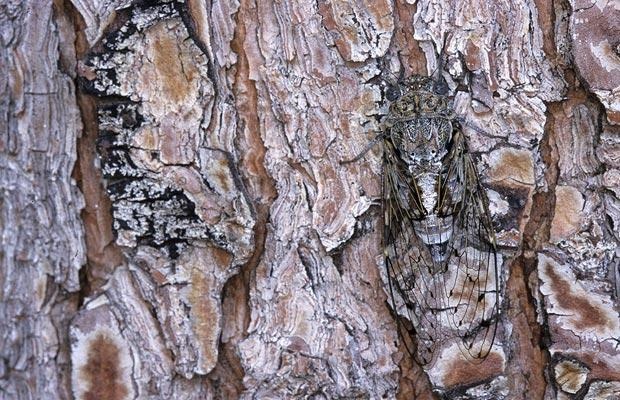}
    \hspace{-3pt} \includegraphics[width=0.14\textwidth,height=0.14\textwidth]{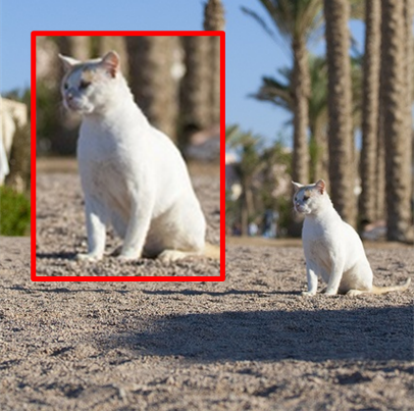}
    \hspace{-3pt} \includegraphics[width=0.14\textwidth,height=0.14\textwidth]{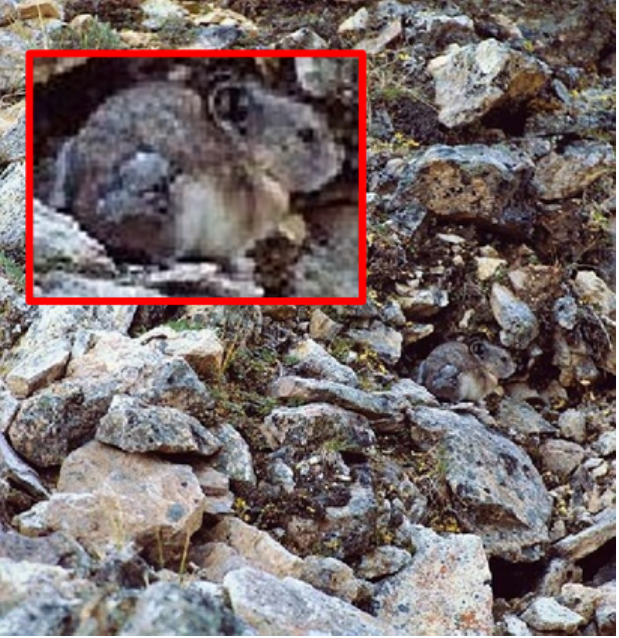}
    \raisebox{1\height}{\makebox[0.05\textwidth]{}}
    \vspace{1pt}
    \\
    \raisebox{3\height}{\makebox[0.14\textwidth]{\Large Mask}}
    \includegraphics[width=0.14\textwidth,height=0.14\textwidth]{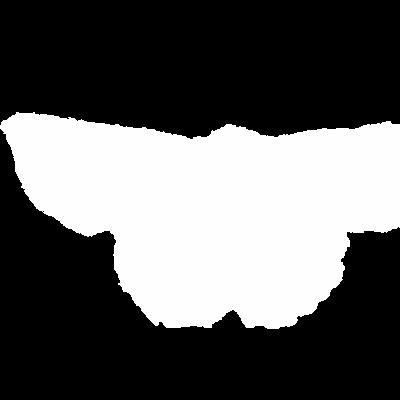}
    \hspace{-3pt} \includegraphics[width=0.14\textwidth,height=0.14\textwidth]{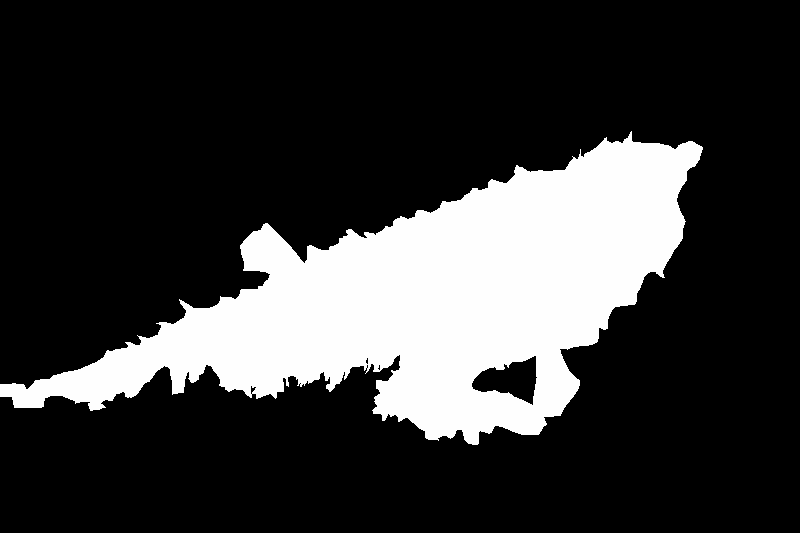}
    \hspace{-3pt} \includegraphics[width=0.14\textwidth,height=0.14\textwidth]{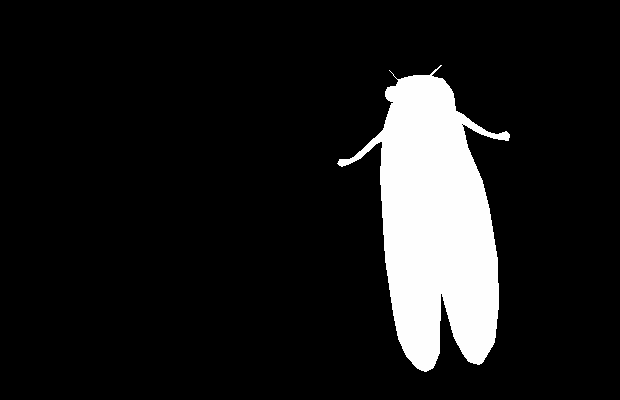}
    \hspace{-3pt} \includegraphics[width=0.14\textwidth,height=0.14\textwidth]{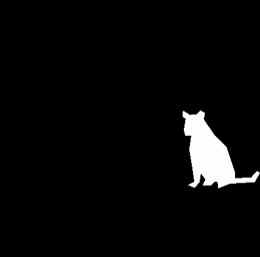}
    \hspace{-3pt} \includegraphics[width=0.14\textwidth,height=0.14\textwidth]{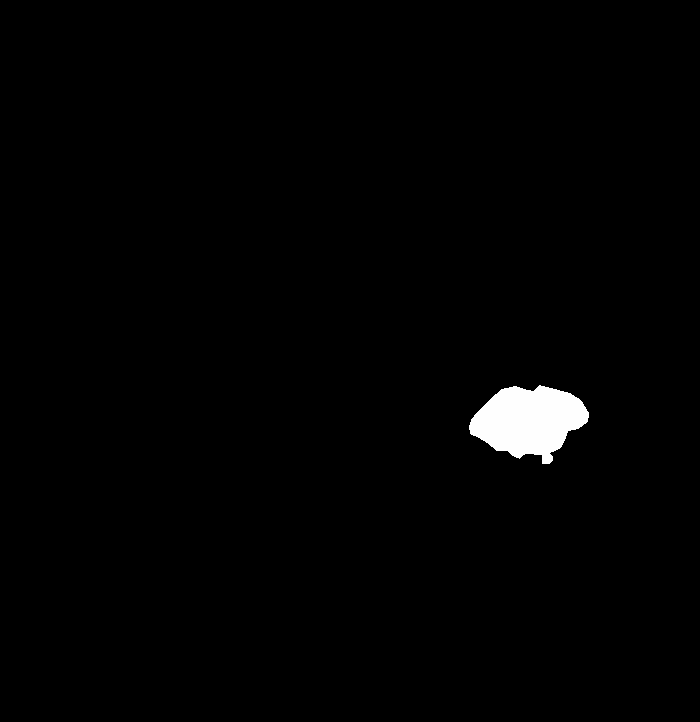}
    \raisebox{1\height}{\makebox[0.05\textwidth]{}}
    \vspace{1pt}
    \\
    \raisebox{3\height}{\makebox[0.14\textwidth]{\Large LAKE-RED}}
    \includegraphics[width=0.14\textwidth,height=0.14\textwidth]{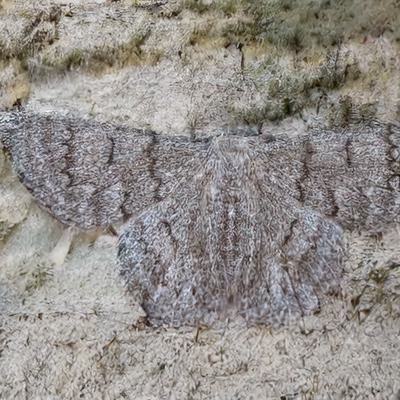}
    \hspace{-3pt} \includegraphics[width=0.14\textwidth,height=0.14\textwidth]{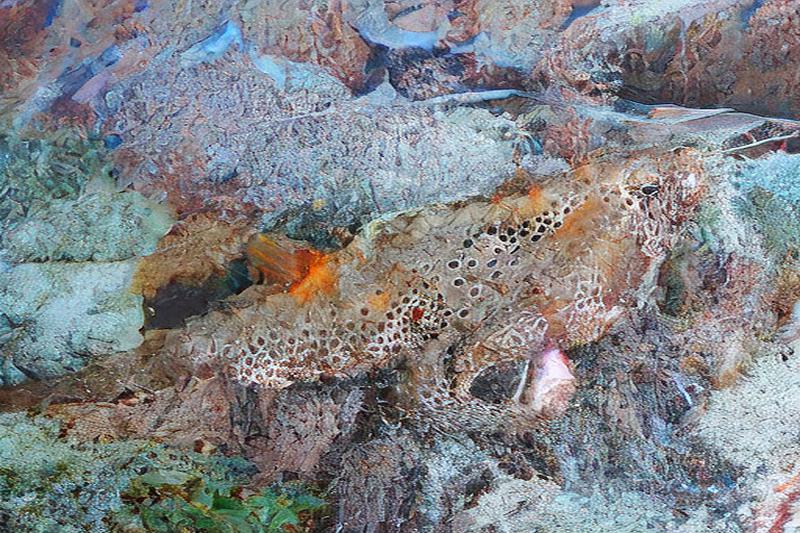}
    \hspace{-3pt} \includegraphics[width=0.14\textwidth,height=0.14\textwidth]{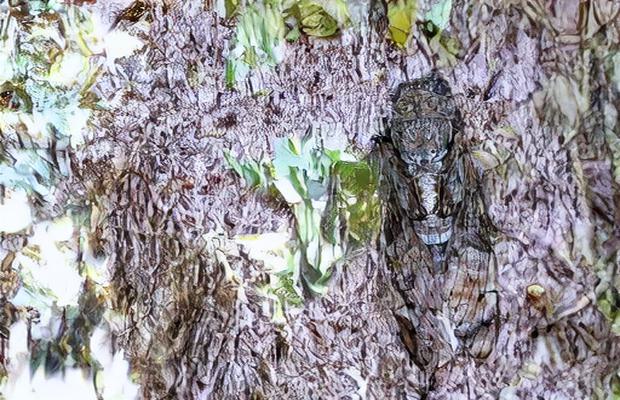}
    \hspace{-3pt} \includegraphics[width=0.14\textwidth,height=0.14\textwidth]{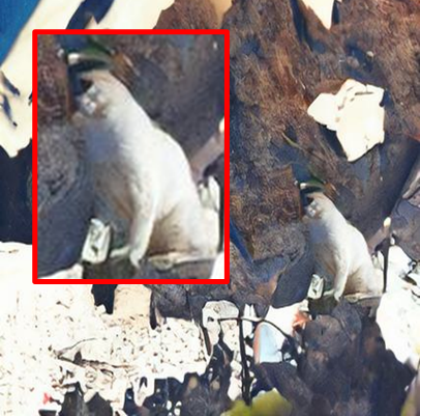}
    \hspace{-3pt} \includegraphics[width=0.14\textwidth,height=0.14\textwidth]{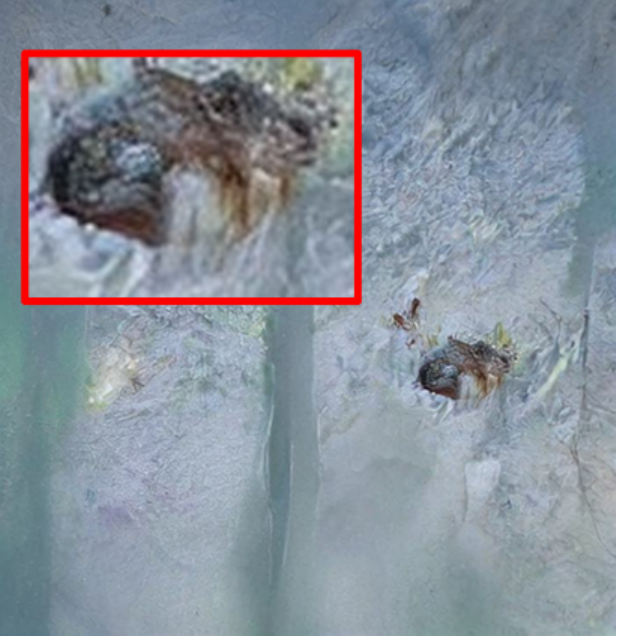}
    \raisebox{1\height}{\makebox[0.05\textwidth]{}}
    \vspace{2pt}
    \\
    \raisebox{3\height}{\makebox[0.14\textwidth]{\Large Ours}}
    \includegraphics[width=0.14\textwidth,height=0.14\textwidth]{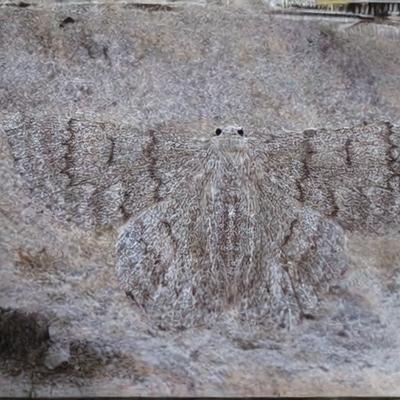}
    \hspace{-3pt} \includegraphics[width=0.14\textwidth,height=0.14\textwidth]{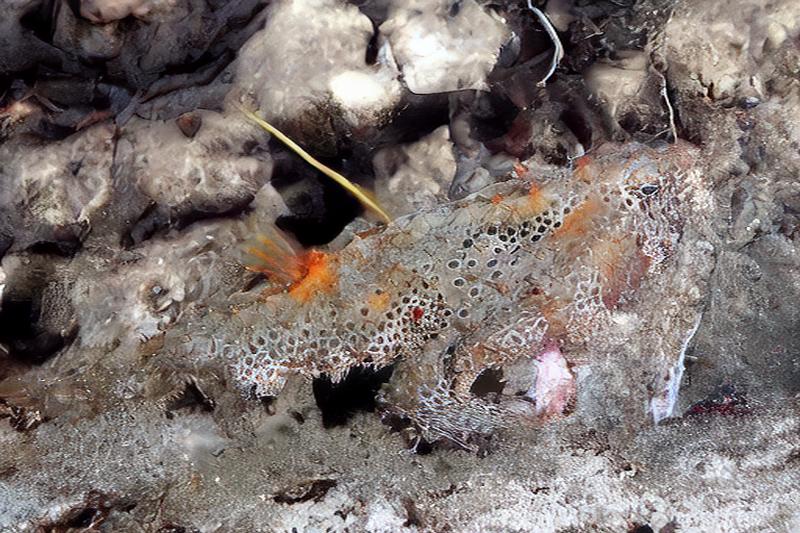}
    \hspace{-3pt} \includegraphics[width=0.14\textwidth,height=0.14\textwidth]{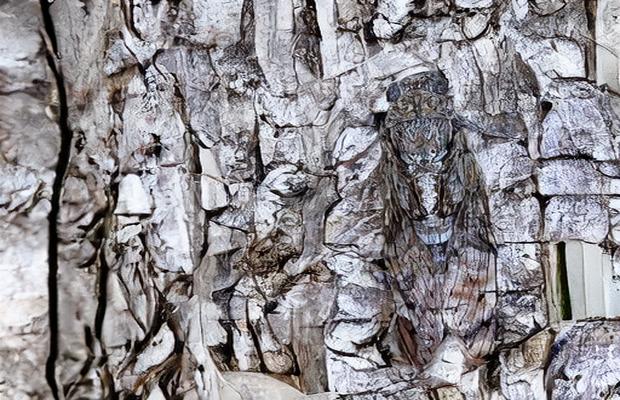}
    \hspace{-3pt} \includegraphics[width=0.14\textwidth,height=0.14\textwidth]{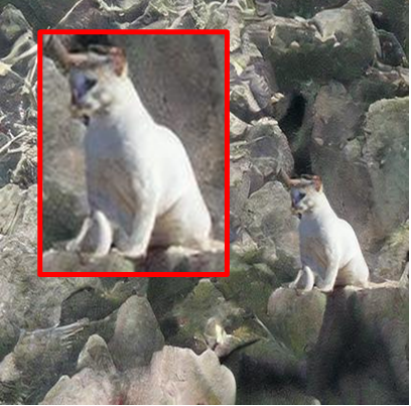}
    \hspace{-3pt} \includegraphics[width=0.14\textwidth,height=0.14\textwidth]{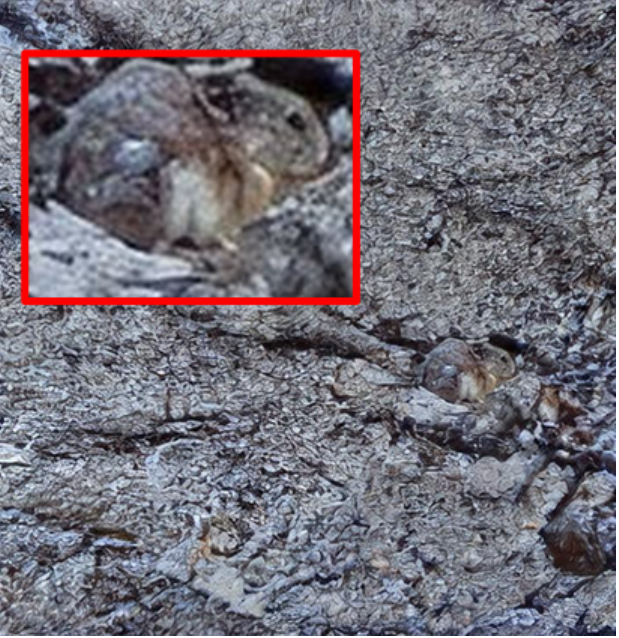}
    \raisebox{1\height}{\makebox[0.05\textwidth]{}}
    \vspace{4pt}
    \\
    \makebox[0.16\textwidth][r]{\large Large}
    \begin{tikzpicture}
    \draw[teal, line width=1.5mm, ->] (9,0) -- (20,0);
    \end{tikzpicture}
    \makebox[0.07\textwidth][l]{\large Small}
    \caption{Qualitative comparison of results for objects of varying sizes, with object size decreasing from left to right. For better visualization of small objects, the foreground is zoomed in for the last two columns (red box)}
    \label{fig:object_size}
\end{figure*}

\section{User Study}
\begin{figure*}[h]
    \centering
    \includegraphics[width=0.8\textwidth]{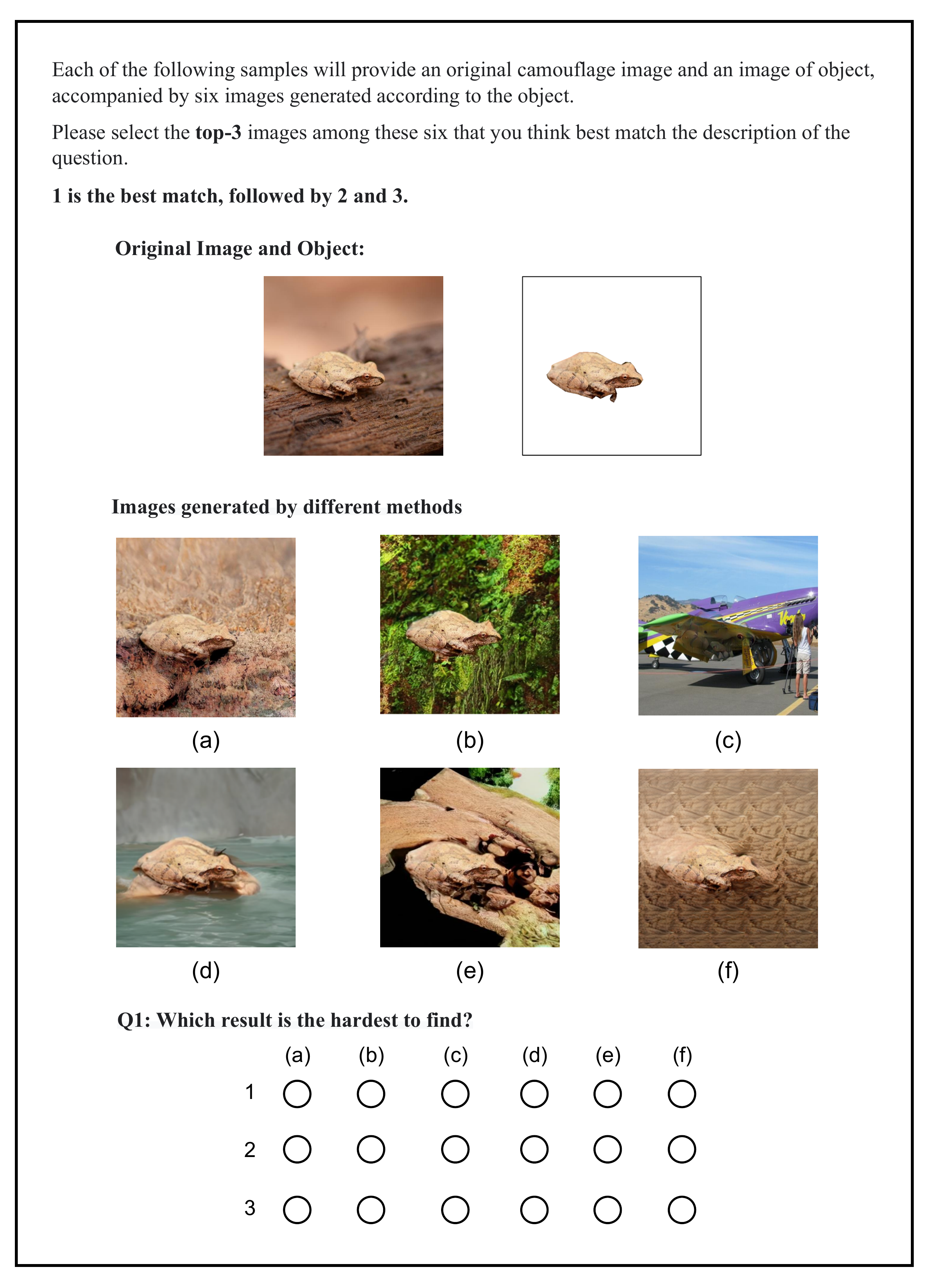}
    \caption{The interface of our user study for human evaluation.}
    \label{fig:userstudy}
\end{figure*}

Fig. \ref{fig:userstudy} shows the interface for human evaluation. Twenty foreground image sets were randomly selected. For each image set, we shuffled the order of the images to prevent biases. Participants were asked to rate the generated results based on three questions, selecting their top 3 choices for each question, with 1 being the highest. Detailed feedback was collected from 22 participants, most of whom have a background in deep learning.

\bibliographystyle{IEEEbib}
\bibliography{supp}